\documentclass[10pt,twocolumn,letterpaper]{article}
\usepackage{lmodern}

\usepackage{cvpr}              % To produce the CAMERA-READY version

\usepackage[accsupp]{axessibility}
\usepackage{newfloat}
\usepackage{listings}

\usepackage{graphicx}
\usepackage[accsupp]{axessibility}
\usepackage{amsmath}
\usepackage{amssymb}
\usepackage{mathtools}
\usepackage{subcaption}

\graphicspath{ {./figs/} }
\usepackage{booktabs}
\usepackage{multirow}
\usepackage{algorithm}
\usepackage{algorithmic}
\usepackage{enumitem}
\usepackage[flushleft]{threeparttable}
\usepackage{color}
\usepackage{xcolor}
\usepackage{colortbl}

\usepackage{diagbox}
\usepackage{tabularx}

\newcommand{\myparagraph}[1]{\noindent\textbf{#1.}}
\newcolumntype{P}[1]{>{\centering\arraybackslash}p{#1}}

\def\ie{i.e.}
\def\eg{e.g.}

\def\and{\textrm{and}}

\def\Diag{\textrm{Diag}}

\def\Diag{\textrm{Diag}}

\def\0{\textbf{0}}
\def\1{\textbf{1}}

\def\s{\boldsymbol{s}}

\def\x{\boldsymbol{x}}
\def\y{\boldsymbol{y}}
\def\z{\boldsymbol{z}}

\def\I{\mathbf{I}}

\def\A{\mathcal{A}}
\def\C{\mathcal{C}}
\def\D{\mathcal{D}}

\def\F{\mathcal{F}}

\def\N{\mathcal{N}}

\def\P{\mathcal{P}}

\def\S{\mathcal{S}}
\def\T{\mathcal{T}}

\def\X{\mathcal{X}}
\def\Y{\mathcal{Y}}
\def\Z{\mathbf{Z}}
\def\RR{\mathbb{R}}

\newcommand{\semismall}{\fontsize{8}{10}\selectfont}

\newcommand{\silentfootnote}[1]{%
  \begingroup% 开启局部作用域
  \renewcommand{\thefootnote}{}% 清空脚注序号
  \footnote{#1}% 添加脚注内容
  \addtocounter{footnote}{-1}% 还原脚注计数器，不影响后续正常脚注
  \endgroup% 结束局部作用域
}

\newcommand{\mcb}{\color{black}}

\usepackage[dvipsnames]{xcolor}
\definecolor{myGray}{gray}{0.9}

\definecolor{cvprblue}{rgb}{0.21,0.49,0.74}
\usepackage[pagebackref,breaklinks,colorlinks,allcolors=cvprblue]{hyperref}

\usepackage[capitalize]{cleveref}

\crefname{section}{Sec.}{Secs.}
\Crefname{section}{Section}{Sections}
\Crefname{table}{Table}{Tables}
\crefname{table}{Tab.}{Tabs.}

\def\confName{CVPR}
\def\confYear{2026}

\title{Multi-Modal Representation Learning via Semi-Supervised Rate Reduction \\ for Generalized Category Discovery}

\author{Wei He$^1$ %$^\dag$
,~Xianghan Meng$^1$ %$^\dag$
,~Zhiyuan Huang$^1$ %$^\dag$
,~Xianbiao Qi$^2$ %$^\ddag$
,~Rong Xiao$^2$ %$^\ddag$
,~Chun-Guang Li$^{\ast, 1}$\\ %$^{\ast,\dag}$\\
$^1$ School of Artificial Intelligence, Beijing University of Posts and Telecommunications \\
{\tt\small \{wei.he, huangzhiyuan, mengxianghan, lichunguang\}@bupt.edu.cn} \\
$^2$ Intellifusion Inc.,{~Shenzhen},{~P.R. China} \\
}

\begin{document}
\maketitle
%{\mcb 
\begin{abstract}
%Generalized Category Discovery (GCD) aims to identify both known and unknown categories, with only partial labels given for the known categories, posing a challenging open-set recognition problem. State-of-the-art approaches for GCD are usually built on multi-modality representation learning, which heavily depends upon inter-modality alignment {\mcb but pays less attention to intra-modality alignment.} {\mcr However, few of them perform a proper intra-modality alignment to generate a desired underlying structure in representation distributions.} In this paper, we propose a novel and effective multi-modal representation learning approach for GCD via Semi-Supervised Rate Reduction, called SSR$^2$-GCD, to learn cross-modality representations with desired structural properties via properly harnessing intra-modality alignment. %by emphasizing the proper alignment of intra-modality relationships. Moreover, to boost knowledge transfer, we integrate prompt candidates by leveraging the inter-modal alignment offered by Vision Language Models. We conduct extensive experiments on generic and fine-grained benchmark datasets, demonstrating the superior performance of our approach.

Generalized Category Discovery (GCD) aims to identify both known and unknown categories, with only partial labels given for the known categories, posing a challenging open-set recognition problem. State-of-the-art approaches for GCD are usually built on multi-modality representation learning, which pays heavily attention to inter-modality alignment rather than intra-modality alignment. 
In this paper, we propose a novel and effective multi-modal representation learning approach for GCD via Semi-Supervised Rate Reduction, called SSR$^2$-GCD, to learn cross-modality representations with desired underlying structures via properly harnessing intra-modality alignment. Moreover, to boost knowledge transfer, we integrate the information from prompt candidates by leveraging the inter-modal alignment offered by Vision Language Models. We conduct extensive experiments on generic and fine-grained benchmark datasets, demonstrating superior performance of the proposed approach and verifying the importance of intra-modality alignment. 
The code is available at: \url{https://github.com/hewei98/SSR2-GCD}.
% harnessing an proper 

\silentfootnote{$^\ast$ Chun-Guang Li is the corresponding author.}

\end{abstract}

\section{Introduction}
\label{sec:intro}
Generalized Category Discovery (GCD) has emerged as a natural and challenging extension of open-set recognition with the aim of discovering categories (\ie, patterns) in the open world~\citep{Vaze:CVPR22-GCD, Scheirer:TPAMI12-OSR}.
The goal of GCD is to recognize both known and unknown categories, going beyond the standard open-set recognition problem by leveraging knowledge learned from known categories to discover unknown categories.
For example, in the typical setting of GCD task, half of categories are partially labeled (known) and the rest of categories remain unlabeled (unknown). 
Such a setting is relevant to real-world exploration scenarios in which data exhibit a mixture of known and unknown categories. %class structures.

Roughly speaking, existing state-of-the-art approaches to address GCD problem %usually 
follow a two-phase framework: a) generating representations for images by fine-tuning the pre-trained models, and b) applying clustering algorithms on the learned representations of all unlabeled data.
%
% However, these approaches lack of effective signals to transfer knowledge from known categories to discovering unknown categories,
%
% that is, almost all of them are using the given partial labels to recognize the known categories but recover the unknown categories in unsupervised or self-supervised manner. 
However, the representations derived from a single visual modality may lack sufficient discriminative information, especially when discovering visually similar categories.

For %our 
human being, it is consciously or unconsciously leveraging information or cues from multiple modalities %sources %{\mcr from known categories} 
to recognize unknown categories. 
Recently, there are a few attempts to explore multi-modal frameworks for GCD in image modality by integrating information from textual modality. 
For example, CLIP-GCD~\citep{Rabha:Arxiv23-clipgcd} leverages a knowledge database to search for similar texts of query images; TextGCD~\citep{Zheng:ECCV24-textgcd} constructs prompts from tag and attribute lexicons; and GET~\citep{Wang:CVPR25-get} learns a textual inversion network to generate prompts.
These frameworks for multi-modal GCD have exploited textual cues into visual datasets to perform inter-modality alignment for cross-modality representation learning, but still lack sufficient consideration on the underlying structure of the distribution in the learned representation.
For example, an existing CLIP-style inter-modality loss~\citep{Radford:ICML2021-CLIP} or a GCD-style intra-modality loss~\citep{Vaze:CVPR22-GCD} 
is incorporated to conduct inter-modality alignment for representation learning, lacking of a proper loss %effective  means 
to harness intra-modality alignment. 
Thus the insights into %the role of 
the interactions between inter-modality alignment and intra-modality alignment in multi-modal representation learning are still vague.

In this paper, we attempt to incorporate the maximal coding rate reduction principle~\citep{Yu:NIPS20} into multi-modal representation learning for GCD, and present a novel and efficient approach, called Semi-Supervised Rate Reduction %(SSR$^2$) for GCD 
(SSR$^2$-GCD).
Specifically, 
in SSR$^2$-GCD, a so-called %Semi-Supervised Rate Reduction (SSR$^2$) 
SSR$^2$ loss, which encourages intra-modal consistency, is employed to learn desired structured representations. 
%in which %encourage 
%the consistency of intra-modal representations is encouraged.
%
Unlike the existing multi-modal GCD approaches, in SSR$^2$-GCD, the learned cross-modality representations for both known and unknown categories are compressed in a balanced manner. 
% equivalently compressed into target structures. %into evenly distributed structure.
% 
% The argue of inter-modal alignment is not that important and thus the sentence below is not needed either.
% This finding is thoroughly discussed and validated through extensive experiments.
% 
Moreover, we also present a Retrieval-based Text Aggregation (RTA) strategy to enhance the text generation, in which the information from %a larger amount of 
augmented prompt candidates is %integrated 
aggregated to generate semantic-rich textual representations.

\myparagraph{Paper Contributions} 
The % main 
contributions of the paper are highlighted as follows. 
\begin{itemize}
%enumerate
\item We propose a Semi-Supervised Rate Reduction framework for GCD %for Generalized Category Discovery 
(SSR$^2$-GCD) to learn structured representations with desired distribution structures. 

\item We present an effective Retrieval-based Text Aggregation (RTA) strategy for generating semantic-rich representations to boost knowledge transfer.

\item We conduct extensive experiments, %on eight datasets, 
showing superior performance of the proposed approach and validating the importance of intra-modality alignment. %the claims on learned representations.
\end{itemize}

\section{Relate Work}
\label{sec:relate}

\myparagraph{Generalized Category Discovery} 
GCD considers a realistic scenario in which the unlabeled dataset includes samples from both known and unknown categories, requiring simultaneous discovering of known and unknown categories.
Initially, \citet{Vaze:CVPR22-GCD} leverage supervised and self-supervised contrastive learning to refine the features produced by pre-trained vision models and use $k$-means~\citep{MacQueen-1967, Arthur:SIAM2007-kmeans++} %and $k$-means++~\citep{Arthur:SIAM2007-kmeans++} 
algorithms for clustering.
%
%A number of methods follow such a pipeline.
%
%For instance, 
Then, SimGCD~\citep{Wen:ICCV23-SimGCD} %which is a notable baseline for GCD, 
introduces a parametric classifier to replace the non-parametric clustering and incorporates an entropy regularization to alleviate the degradation of classifier on unknown categories;
SelEx~\citep{Rastegar:ECCV24-SelEX} leverages a hierarchical semi-supervised $k$-means and achieves better results on fine-grained datasets; 
GPC~\citep{Zhao:ICCV23-GPC} employs a mixture of Gaussian %mixture 
models to learn representations while %simultaneously 
estimating the number of unknown categories; 
PromptCAL~\citep{Zhang:CVPR2023-promptCAL} introduces a contrastive affinity learning framework, in which auxiliary visual prompts are incorporated to address the false negative-induced category collision issue; 
%
% SPTNet~\citep{Wang:ICLR2024-SPTNet} proposes a spatial prompt tuning method that iteratively fine-tunes the backbone and learns pixel-level prompts, effectively transferring semantic knowledge in GCD; 
%
HypCD~\citep{Liu:CVPR25-HypCD} considers both hyperbolic distance and %the 
angle between samples to learn hierarchy-aware representations.
However, all these approaches %methods 
mentioned above exploit the visual cues only.
% 
% Note that for human being, each known visual category inherently corresponds to specific semantic meanings described by natural language and thus it is typically incorporating cues from multiple modalities to recognize new categories.
%

\myparagraph{Multi-modal Generalized Category Discovery}
% Vision-language pre-trained models (VLMs) such as CLIP~\citep{Radford:ICML2021-CLIP} embed images and text into an aligned semantic space by pulling the representations of positive image-text pairs, enabling various downstream applications.
%
%Recently, 
Existing Multi-modal GCD methods usually leverage the external guidance of textual modality brought by Vision-language pre-trained models (VLMs) to facilitate knowledge transfer between known and unknown categories.
For instance, CLIP-GCD~\citep{Rabha:Arxiv23-clipgcd} leverages a knowledge database to generate texture descriptions and concatenates both the visual embedding and text embedding obtained from a frozen CLIP backbone for clustering; 
%
% MM-GCD~\citep{Su:ArXiv24-MMGCD} propose a multi-modal framework to align with both the feature and output spaces of different modalities using contrastive learning and distillation technique; 
%
TextGCD~\citep{Zheng:ECCV24-textgcd} proposes a retrieval-based text generation method to generate semantic-rich texture descriptions by incorporating abundant tag and attribute candidates and introduces a co-teaching technique to align the clustering outputs of vision and text branches.
However, TextGCD simply employs the inter-modal contrastive loss of CLIP to fine-tune the backbone and %while using 
uses the similarities within each modality for intra-modal clustering.
%
%More 
Recently, 
GET~\citep{Wang:CVPR25-get} trains a textual inversion network~\citep{Baldrati:ICCV23-OTI} that maps the image embedding to pseudo textual token for unlabeled images.
%
% To alleviate the distributional shift of the text manifold towards the image manifold, GET aligns the embedding of pseudo-token with the embedding of class names on labeled data.
%
In the GCD setting, however, only class names of known categories are available %, which prevents 
preventing the textual inversion network from generating high-quality pseudo-tokens for images belonging to unknown categories.
{\mcb 
%Although %While 
%these %existing multi-modal GCD 
%approaches have achieved 
%Moreover, while 
Although these methods achieve promising results, %they still fail to fully exploit the information provided by the textual modality. 
%the insights into 
the interaction between inter-modal alignment and intra-modal alignment is unclear yet. }

\section{Preliminaries}
%In this section, we review the setting of GCD problem and the multi-modal GCD pipelines relevant to our approach.

\subsection{Problem Notation}
Denote data set as $\D_L \cup \D_U$, %which consists 
consisting of labeled data $\D_L = {(\x_i, y_i^\ell)}_{i=1}^{M} \subseteq \X \times \Y_L $ and unlabeled data $\D_U = {(\x_i, y_i^u)}_{i=1}^{N} \subseteq \X \times \Y_U$, where $\Y_L$ and $\Y_U$ denote the label sets %spaces, 
and $\Y_L\subset \Y_U$.
In GCD setting, the labeled samples in $\D_L$ are from the known categories; whereas unlabeled samples in $\D_U$ are either from the known categories %nor 
or from some unknown categories. 
As usual, the total number of categories $K=|\Y_U|$ is given (or can be estimated).
The goal of GCD is to estimate the labels of samples in $\D_U$.

\subsection{Multi-modal GCD Pipelines} 
%
%The common practice of multi-modal GCD frameworks consist of three components: text generation, representation learning, and clustering.
The common multi-modal GCD 
pipelines consist of three components: a) text generation, b) representation learning and c) clustering.

\myparagraph{Text Generation}
To introduce textual cues, %to visual datasets, %inherent inter-modality alignment properties of 
pre-trained VLMs are usually leveraged to generate pseudo-texts for query images, in which  
%
%Among these approaches, 
retrieval-based methods~\citep{Li:ICML24-TAC,Zheng:ECCV24-textgcd} are used to construct a prompt %database 
set $\P$ and search for the suitable %optimal 
prompt $p \in \P$ for the query image. 
%that maximizes the cosine similarity between the embeddings of query image and prompt candidates.
%
For instance, TextGCD~\citep{Zheng:ECCV24-textgcd} uses the %class 
category name in ImageNet~\citep{Deng:CVPR2009-imagenet} to construct a tag lexicon and leverages GPT3~\citep{Brown:NIPS20-GPT} to generate %distinguishing 
attributes for these tags, in which the tags and attributes are used to construct the prompt for each query image.
%
%For each query image, the top-3 similar tags and top-2 similar attributes are used to construct the prompt: 
%
%``most likely $\{\text{tag}_1\}$, perhaps $\{\text{tag}_2\}$, likely $\{\text{tag}_3\}$, most likely $\{\text{attr}_1\}$, perhaps $\{\text{attr}_2\}$''.

\myparagraph{Representation Learning}
Given query images and their corresponding pseudo-texts, multi-modal GCD frameworks usually further refine %their 
the representations. %simultaneously.
For instance, TextGCD~\citep{Zheng:ECCV24-textgcd} learns the representations by simply using a CLIP-style inter-modal contrastive loss % \ie, 
\begin{equation}
    \label{eq:clip_inter_contrast}
    \mathcal{L}_\text{CLIP} = -\frac{1}{|B|} \sum_{i \in B} \log \frac{\exp \left( {\z_i^{\text{I}}}^\top \z_i^{\text{T}} / \tau_c \right)}{\sum_{j\neq i} \exp \left( {\z_i^{\text{I}}}^\top \z_j^{\text{T}} /\tau_c \right)},
\end{equation}
where $B$ denotes the mini-batch of data, $\z_i^{\text{I}},\z_i^{\text{T}}\in\mathbb R^{d}$ denote the embeddings of the $i$-th image and pseudo-text, and $\tau_c$ is the temperature factor.
Minimizing %Optimizing 
$\mathcal{L}_\text{CLIP}$ encourages the inter-modal alignment between image and text, % manifolds, 
but does not account for intra-modal alignment within each modality.
GET~\citep{Wang:CVPR25-get} encourages the inter-modal alignment in representation learning, while incorporating the %widely-used 
following supervised and unsupervised contrastive losses %to encourage the intra-modal alignment, %align intra-modal relationships within each modality, \ie,
\begin{align}
    \label{eq:con_rep_loss}
    %\begin{split}
    & \mathcal{L}_\text{con}=\lambda\mathcal{L}_\text{con}^\text{s}+(1-\lambda)\mathcal{L}_\text{con}^\text{u}, \\ %\nonumber \\
    & \mathcal{L}_\text{con}^\text{s} = - \sum_{i \in B_\ell} \frac{1}{|\mathcal{N}_i|} \sum_{j \in \mathcal{N}_i} \log \frac{\exp \left( \boldsymbol{z}_i^\top \boldsymbol{z}_j'/ \tau_a \right)}{\sum_{ m \neq i} \exp \left( \boldsymbol{z}_i^\top \boldsymbol{z}_m'/ \tau_a \right)}, \nonumber \\
    & \mathcal{L}_\text{con}^\text{u} = - \sum_{i \in B} \log \frac{\exp \left( \boldsymbol{z}_i^\top \boldsymbol{z}_i'/\tau_b \right)}{\sum_{ m \neq i} \exp \left( \boldsymbol{z}_i^\top \boldsymbol{z}_m'/\tau_b \right)}, \nonumber %\\
    %\end{split}
\end{align}
to encourage the intra-modal alignment, where $\lambda$ is the balancing parameter, $B_\ell$ denotes the labeled subset of $B$, $\boldsymbol{z}_i$ and $\boldsymbol{z}_i'$ are embeddings of augmented data pairs, $\N_i$ is the index set of samples %data indices 
with the same label as the $i$-th sample, %data, 
and $\tau_a$ and $\tau_b$ are the temperature parameters. 
Still, the intra-modal representation learning in multi-modal GCD frameworks adheres to the paradigm of uni-modal counterparts, and fails to resolve its inherent issue, \ie, %problem. 
%
%That is, 
minimizing %optimizing 
$\mathcal{L}_\text{con}$ results in an \emph{imbalanced compression} of embeddings, %since
because $\mathcal{L}_\text{con}^\text{u}$ pulls positive augmented data pairs across all categories, whereas $\mathcal{L}_\text{con}^\text{s}$ further pulls labeled data only for known categories.
As illustrated in Figure~\ref{fig:overview}, such an imbalanced compression issue prevents clustering algorithms from identifying clusters accurately. % boundaries.

\myparagraph{Clustering}
Existing multi-modal GCD frameworks usually 
follow SimGCD~\citep{Wen:ICCV23-SimGCD} to cluster unlabeled data %to cluster unlabeled data, which consists of 
%usually a supervised cross-entropy loss~\citep{Kri:NIPS12}, an unsupervised self-distillation loss~\citep{Assran:ECCV22-distill} and an entropy regularizer are integrated to cluster unlabeled data, \ie, }
%usually 
via integrating a cross-entropy loss~\citep{Kri:NIPS12}, a self-distillation loss~\citep{Assran:ECCV22-distill} and an entropy regularizer, %are integrated to cluster unlabeled data, 
\ie, 
\begin{equation}
    \label{eq:detailed_simgcd_classifier}
    \mathcal{L}_\text{cls} =\sum_{i\in B_\ell} \ell_\text{CE}(\y_i^*,\y_i) + \gamma \sum_{i\in B} \ell_\text{CE}({ \y}^\prime_{i},\y_i)-\mu H({\bar{\y}}),
\end{equation}
where $\gamma$ and $\mu$ are the balancing parameters, $\ell_\text{CE}$ denotes the cross-entropy loss, $\y_i^*$ is the ground-truth label of the $i$-th image, $\y_i$ is the prediction of the $i$-th embedding $\z_i$, $ \y_i^\prime$ is the prediction of the augmented counterpart $\z'_i$ with a sharper temperature in \texttt{softmax}, and %The mean 
an entropy regularizer $H(\bar{\y})=-\sum_k \bar{\y}^{(k)} \log \bar{\y}^{(k)}$ is used %introduced 
to prevent degenerated predictions in new classes, in which %where 
$\bar{\y} = \frac{1}{2|B|}\sum_{i\in B}(\y_i+\y'_i)$ denotes the mean prediction of $\y_i+\y'_i$ in a mini-batch using the same temperature, and $\bar{\y}^{(k)}$ is the entry of $\bar{\y}$ associating to the $k$-th class. 
For example, GET trains an %single 
MLP by %optimizing 
the loss in~Eq.~\eqref{eq:detailed_simgcd_classifier} to produce predictions for both image and text embeddings, whereas TextGCD implements dual-branch classifiers to handle image and text embeddings, respectively, and 
%
%Additionally, TextGCD 
%uses %the 
%a co-teaching strategy in which high-confidence samples are used to supervise the %learning of 
%classifiers training, 
use high confidence samples to supervise the classifier training via a co-teaching loss
%\ie,
\begin{equation}
    \label{eq:co_teach}
    \mathcal{L}_\text{co-teach}=  \sum_{i \in \S^\text{I}} \ell_\text{CE}(\y_{i}^{\mathrm{I}}, \boldsymbol{y}_{i}^{\mathrm{T}}) + \sum_{j \in \S^\text{T}} \ell_\text{CE}(\y_{j}^{\mathrm{T}}, \boldsymbol{y}_{j}^{\mathrm{I}}),
\end{equation}
where $\S^\text{I}$ and $\S^\text{T}$ are the sample sets %that are 
selected based on the confidence score of $\y^\text{I}$ and $\y^\text{T}$, respectively.

\section{Our Proposed Approach: SSR$^2$-GCD}
\label{sec:method}

\begin{figure}[t]
\vskip -4pt
    \centering
    \includegraphics[width=0.96\linewidth]{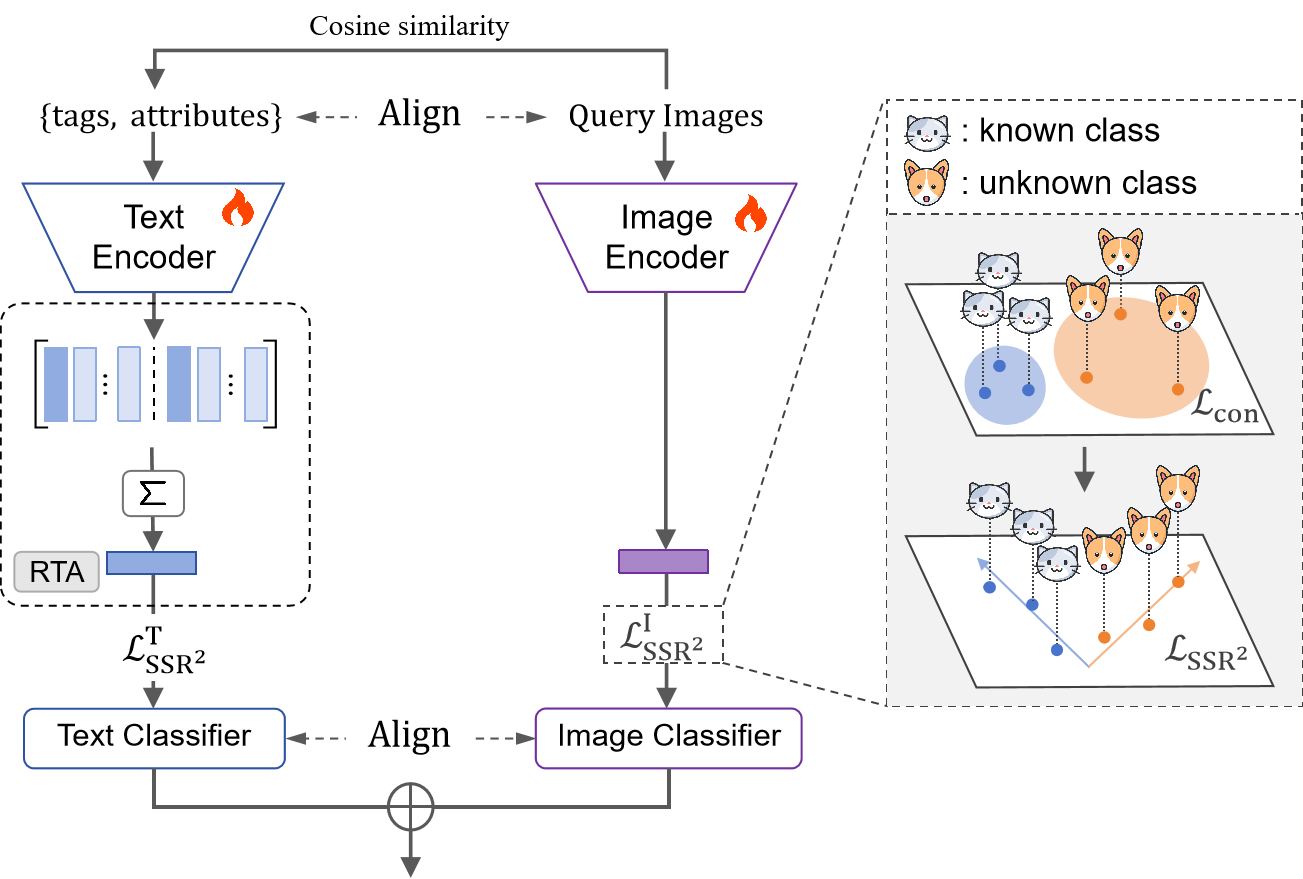}
    \caption{Illustration of our proposed framework: SSR$^2$-GCD.}
    \label{fig:overview}
\vskip -4pt
\end{figure}

Our %proposed framework 
SSR$^2$-GCD consists of three of modules: a) Retrieval-based Text Aggregation (RTA) module for text generation and aggregation; 
%for aggregating the embeddings of prompts to incorporate the textual information; 
b) Semi-Supervised Rate Reduction (SSR$^2$) module for representation learning; and c) dual-branch classifiers to learn pseudo-labels. % from each modality. 
%
%Specifically, at first %we use 
%RTA is used to aggregate the embeddings of prompts to incorporate the textual information; %that is helpful in discovering unknown categories; 
%
%then %we use Semi-Supervised Rate Reduction (SSR$^2$) 
%SSR$^2$ modules are used to learn structured representations, and finally 
%
%the dual-branch classifier is deployed to learn pseudo-labels from each modality.
%
For clarity, we illustrate our proposed framework in Figure~\ref{fig:overview}.

\subsection{Retrieval-based Text Aggregation}
\label{subsec:text_generation}

% For text generation, we follow TextGCD~\citep{Zheng:ECCV24-textgcd} to construct tag and attribute lexicons, and then search for similar tags and attributes, since that incorporating more information from multiple tag and attribute candidates is helpful to discover the unknown categories. }
Since that incorporating more information from multiple tag and attribute candidates is helpful to discover the unknown categories, we follow TextGCD~\citep{Zheng:ECCV24-textgcd} for text generation, \ie, constructing tag and attribute lexicons and then searching for similar tags and attributes. 
%
% 
%Still, due to CLIP's limitation in handling long textual prompts, the way of constructing prompts in TextGCD is sub-optimal, because that CLIP fails to generate satisfactory embeddings for prompts exceeding 20 tokens~\citep{Zhang:ECCV24-longclip}.
%
However, the way to construct prompts in TextGCD is sub-optimal because that CLIP fails to generate satisfactory embeddings for prompts exceeding 20 tokens due to its limitation in handling long textual prompts~\citep{Zhang:ECCV24-longclip}. 
Given CLIP text encoder $\F^\text{T}$ and tokenizer $\T$, instead, in this paper, we use the text encoder $\F^\text{T}$ to embed the prompts of the top-$c$ most similar tag and attribute and %then 
compute the textual embedding by:
\begin{equation}
    \label{eq:RTA}
    \z^{\text{T}}= \sum_{i=1}^c \sigma_i [\F^{\text{T}}( \T (a_i)) +  \F^{\text{T}}(\T (b_i))] ,
    %\z^{\text{T}}= \sum_{i=1}^c \sigma_i \F^{\text{T}}( \T (a_i)) + \sum_{i=1}^c \sigma_i \F^{\text{T}}(\T (b_i)) ,
\end{equation}
where $a_i$ and $b_i$ are the $i$-th most similar tag and attribute of the query image, respectively, as ranked by the cosine similarity between their embeddings, 
%and the %embedding of query image $z^\text{I}$'s,  
%
% $c$ means that only the top-$c$ most similar tags and attributes are considered,
%
and %the variable 
$\sigma_i$ is defined as %used to assign higher weights to the most similar tag and attribute and aggregate information from other candidates, \ie, 
\begin{equation}
    \sigma_i = 
    \begin{cases} 
    1 - \alpha & \text{if } i = 1 \\
    \frac{\alpha}{c-1} & \text{otherwise},
    \end{cases}
\end{equation}
which is used to assign higher weights to the most similar tag and attribute and aggregate information from other candidates, where $\alpha>0$ is hyper-parameter (\eg, $\alpha=0.5$). 
Such a strategy allows us to aggregate richer information from a larger set of candidates (\eg, $c=4$).

\subsection{Semi-Supervised Rate Reduction Module for Representation Learning}
\label{subsec:representation_learning}

%
%In this section, we tackle the imbalanced compression issue in existing methods.
%
% Specifically, we propose a \textbf{S}emi-\textbf{S}upervised \textbf{R}ate \textbf{R}eduction (\textbf{SSR$^2$}) approach to learn structured representations from intra-modal relationships while achieving even compression across known and unknown categories.
%
To learn structured representations from intra-modal relationships and achieve balanced compression across both known and unknown categories, 
%
%inspired by the structured representation learning principle---Maximal Coding Rate Reduction~\citep{Yu:NIPS20} which is %a structured representation learning technique originally 
%developed for supervised settings, 
we propose a \textbf{S}emi-\textbf{S}upervised \textbf{R}ate \textbf{R}eduction (\textbf{SSR$^2$}) approach, which employs
%we formulate 
a SSR$^2$ objective as follows:
\begin{equation}
\label{eq:our_ssr_rep_learning}
    \mathcal{L}_\text{SSR$^2$} = -R(\Z) + R_c^\text{s}(\Z,\mathbf Y^*) + R_c^\text{u}(\Z,\mathbf Y), %~~~\text{where}
\end{equation}
where
\begin{equation*}
\begin{split}
    & R(\Z) = \log\det \left({\mathbf I} + \frac{d}{N\epsilon^2} \Z \Z^\top \right), \\
    & R_c^\text{s} (\Z_\text{s}, {\mathbf Y^*}) =  \sum_{j=1}^k \frac{N_j^\text{s}}{N} \log\det \left(\mathbf I + \frac{d}{N_j^\text{s}\epsilon^2} \Z_\text{s} \Diag (\mathbf Y_j^*) \Z_\text{s}^\top \right), \\
    & R_c^\text{u} (\Z_\text{u}, {\mathbf Y}) =  \sum_{j=1}^k \frac{N_j^\text{u}}{N} \log\det \left(\mathbf I + \frac{d}{N_j^\text{u}\epsilon^2} \Z_\text{u} \Diag (\mathbf Y_j) \Z_\text{u}^\top \right),
\end{split}
\end{equation*}
where $\Z$ is %denotes 
the embeddings of a mini-batch, $\Z_\text{s}, \Z_\text{u}$ are the embeddings of labeled and unlabeled data, $\mathbf{Y}^*$ is %denotes 
the ground-truth labels, $\mathbf{Y}$ is %denotes 
the pseudo-labels predicted by classifiers, $\I$ is the identity %unit 
matrix, $k$ is the number of categories, $\epsilon>0$ is the hyper-parameter,  %($\epsilon=0.5$ in experiments), 
$N^\text{s}_j$ is the number of labeled data points assigned by $\mathbf{Y}^*$ that belong to the $j$-th class, and $N^\text{u}_j$ is the number of unlabeled data points assigned by $\mathbf{Y}$ that belong to the $j$-th class.

Specifically, during training we fine-tune the image and text encoders of CLIP~\citep{Radford:ICML2021-CLIP} with %, and the loss for each encoder is analogous to $\mathcal{L}_\text{SSR$^2$}$, with the replacement of embedding $\Z$ and pseudo-labels $\mathbf{Y}$ by image and text embeddings $\Z^\text{I}, \Z^\text{T}$ and pseudo-labels of dual-branch classifiers $\mathbf Y^\text{I}, \mathbf Y^\text{T}$, \ie,
the associated loss, %for each encoder, 
that is to replace the embedding $\Z$ and the pseudo-labels $\mathbf{Y}$ by the image and text embeddings $\{\Z^\text{I}, \Z^\text{T}\}$ and the pseudo-labels of dual-branch classifiers $\{\mathbf Y^\text{I}, \mathbf Y^\text{T}\}$, respectively, \ie,
\begin{equation}
\label{eq:ssr_each_branch}
    \begin{split}
        & \mathcal{L}^\text{I}_\text{SSR$^2$} = -R(\Z^\text{I}) + R_c^\text{s}(\Z^\text{I},\mathbf Y^*) + R_c^\text{u}(\Z^\text{I},\mathbf Y^\text{I}), \\
        & \mathcal{L}^\text{T}_\text{SSR$^2$} = -R(\Z^\text{T}) + R_c^\text{s}(\Z^\text{T},\mathbf Y^*) + R_c^\text{u}(\Z^\text{T},\mathbf Y^\text{T}).
    \end{split}
\end{equation}

\myparagraph{Remarks 1}
Our \textbf{SSR$^2$} is inspired by the structured representation learning principle---Maximal Coding Rate Reduction~\citep{Yu:NIPS20}, which is %a structured representation learning technique originally 
developed for supervised settings. % but ours is in semi-supervised setting. 
The effects of introducing SSR$^2$ are twofold. 
%
%On one hand, 
First, maximizing the term $R(\cdot)$ in Eq.~\eqref{eq:our_ssr_rep_learning} expands the entire embeddings, whereas % as a whole, while 
minimizing the terms $R_c^\text{s} (\cdot)$ and $R_c^\text{u} (\cdot)$ encourages the embeddings of each category to span low-dimensional subspaces with %similar 
balanced matrix ranks, 
as proved in~\citep{Yu:NIPS20,Wang:ICML24-Analysis}.
Owing to such a desired property, the representations of both known and unknown categories are \textbf{%equivalently 
compressed in balanced manner}.
To our knowledge, this is the first work to address the imbalanced compression issue in contrastive-based representation learning for GCD. % and to rethink the necessity of inter-modal alignment in the context of multi-modal GCD.
%
% 
%In addition, 
%On the other hand, 
Second, we found in experiments %find 
that the inter-modal alignment leads to \textbf{intra-modal misalignment}; whereas % and can be unnecessary in multi-modal GCD frameworks, as discussed and validated in our experiments.
our SSR$^2$ emphasizes %focuses 
on aligning intra-modal relationships while %accommodating 
tolerating the discrepancies between modalities.

% \subsection{Dual-Branch Clustering}
\subsection{Training Strategy}

\label{subsec:classifier_learning}
To fully discover the differences between modalities, we deploy dual-branch classifiers to separately tackle with image and text embeddings, which are trained %the dual-branch classifiers 
concurrently with the representation learning. 
%
% The training of the two classifiers is conducted concurrently with the representation learning. 
% 
%Finally, 
Specifically, the procedure to train our SSR$^2$-GCD consists of the following two stages.

\myparagraph{Warm-up Stage}
%
%We adopt the loss $\mathcal{L}_\text{cls}$ in Eq.~(\ref{eq:detailed_simgcd_classifier}) for training classifiers.
%
% Specifically, the dual-branch classifiers take the image and text embeddings as the input, yielding modality-specific clustering losses $\mathcal{L}_\text{cls}^\text{I}$ and $\mathcal{L}_\text{cls}^\text{T}$.
%
%Additionally, to 
To prevent erroneous guidance of representation learning induced by pseudo-labels from untrained classifiers,
the self-supervised term $R_c^\text{u}(\Z,\mathbf Y)$ in Eq.~\eqref{eq:our_ssr_rep_learning} is excluded %during
in the early training phase, \ie, $\mathcal{L}_\text{SSR$^2$}$ reduces to 
\begin{equation}
\label{eq:ssr2_warmup}
    \mathcal{L}_\text{SSR$^2$-sup} = -R\left( \Z \right) + R_{c}^{\text{s}}(\Z,\mathbf Y^*).
\end{equation}
To train the classifiers, the loss $\mathcal{L}_\text{cls}$ in Eq.~(\ref{eq:detailed_simgcd_classifier}) is adopted.
%
%Finally, 
Thus the objective in the warm-up stage is: % formulated as:
\begin{equation}
\label{eq:total_warmup_obj}
\mathcal{L}_\text{warm}=\mathcal{L}^\text{I}_\text{SSR$^2$-sup} + \mathcal{L}^\text{T}_\text{SSR$^2$-sup} + \mathcal{L}^\text{I}_\text{cls} + \mathcal{L}^\text{T}_\text{cls},
\end{equation}
where the superscripts $^\text{I}$ and $^\text{T}$ denote the losses applied to %both 
visual and textual modalities, respectively.
%
%where $\mathcal{L}_\text{SSR$^2$-sup}^\text{I}$, $\mathcal{L}_\text{SSR$^2$-sup}^\text{T}$, $\mathcal{L}_\text{cls}^\text{I}$ and $\mathcal{L}_\text{cls}^\text{T}$ denote the losses $\mathcal{L}_\text{SSR$^2$-sup}$ and $\mathcal{L}_\text{cls}$ applied to %both 
%visual and textual modalities, respectively.

\myparagraph{Alignment Stage}
To align the orders of pseudo-labels predicted by dual-branch classifiers, we follow the co-teaching strategy in TextGCD~\citep{Zheng:ECCV24-textgcd}.
Thus by combining the co-teaching loss $\mathcal{L}_\text{co-teach}$ in Eq~\eqref{eq:co_teach}, the total training loss in the alignment stage is: % formulated as:
\begin{equation}
    \label{eq:classifier_align_obj}
    \mathcal{L}_\text{align}=\mathcal{L}^\text{I}_\text{SSR$^2$} + \mathcal{L}^\text{T}_\text{SSR$^2$} +  \mathcal{L}^\text{I}_\text{cls} + \mathcal{L}^\text{T}_\text{cls} +\mathcal{L}_\text{co-teach}.
\end{equation}
After training, the predicted pseudo-label of the $i$-th image is %calculated 
determined by $\texttt{argmax}(\y_i^\text{I}+\y_i^\text{T})$.

\myparagraph{Remarks 2}
The clustering algorithm of our framework is the same as that in TextGCD~\citep{Zheng:ECCV24-textgcd}. % previous multi-modal GCD methods.
However, the key difference lies in representation learning.
Specifically, TextGCD emphasizes only on the inter-modal alignment without incorporating the intra-modal constraints, and the learning of the two classifiers is based on intra-modal similarities.
As pointed out in~\citep{Mistretta:ICLR25-cross}, using inter-modal alignment loss without intra-modal constraint will lead to intra-modal misalignment, \ie, the intra-modal similarities might not correspond to the actual pair-wise relationships, thereby degrading the clustering performance. % performance of intra-modal clustering.
In contrast, in our SSR$^2$-GCD, the intra-modal relationships are well aligned by using $\mathcal{L}_\text{SSR$^2$}$, and thus the two % dual-branch 
classifiers produce satisfactory results.
%
%Additionally, 
In addition, the predictions produced by the dual-branch classifiers are also used as self-supervised signals %that 
to guide the joint representation learning, as shown in Eq.~\eqref{eq:our_ssr_rep_learning}.

\section{Experiments}
\label{sec:experiments}

To validate the effectiveness of our proposed approach, we conduct extensive experiments on eight benchmark datasets and also provide in-depth evaluation and ablation studies.

\myparagraph{Datasets}
We evaluate the performance of GCD methods on four generic datasets, \ie, CIFAR-10/-100~\citep{Krizhevsky:CIFAR2009}, and ImageNet-100/-1k~\citep{Deng:CVPR2009-imagenet}, 
and four fine-grained datasets, \ie, CUB-200-2011~\citep{Wah:2011-cub}, Stanford Cars~\citep{Krause:CVW2013-scars}, Oxford Pets~\citep{Parkhi:CVPR2012-pets} and Flowers102~\citep{Data:Flowers}.
Following the GCD protocol~\citep{Vaze:CVPR22-GCD}, half of the samples from the known categories are selected to form the labeled dataset $\D_L$, while the remaining samples are to form the unlabeled dataset $\D_U$.

\myparagraph{Metrics}
Given the ground-truths of $\D_U$, one can compute the clustering accuracy (ACC) using Hungarian matching algorithm~\citep{kuhn:1955-hungarian}.
Following the GCD protocol~\citep{Vaze:CVPR22-GCD}, we report ACC on all categories (``All''), on known categories (``Old''), and on unknown categories (``New''), respectively.
The average ACC over 3 trials is reported.

\myparagraph{Implementation Details}
In Retrieval-based Text Aggregation, we use CLIP-H/14~\citep{Radford:ICML2021-CLIP} to search prompt candidates to ensure a fair comparison to TextGCD~\citep{Zheng:ECCV24-textgcd}.
During training, we use CLIP-B/16 encoders to produce text and image features.
We report the performance of uni-modal counterparts using DINO~\citep{Caron:ICCV2021-DINO} with ViT-B/16.
The classifier parameters are set with %as in 
the default configurations of SimGCD~\citep{Wen:ICCV23-SimGCD} and TextGCD~\citep{Zheng:ECCV24-textgcd}. %In experiments, 
We use $\epsilon=0.5$ in our SSR$^2$-GCD.
%
% More details are provided in the supplementary.

\label{subsec:sota}
\begin{table}[t]
\centering
\small
\caption{The average ACC (\%) on ImageNet datasets. ``$^\dagger$'' denotes the reproduced results using CLIP backbone.}
\vskip -4pt
\label{tab:sota_2}
\setlength{\tabcolsep}{1.5mm}{
\begin{tabular}{ll ccc ccc}
\toprule
& & \multicolumn{3}{c}{ImageNet-100} & \multicolumn{3}{c}{ImageNet-1k} \\
\cmidrule(lr){3-5} \cmidrule(lr){6-8} 
\multicolumn{2}{c}{Method} & All & Old & New & All & Old & New \\
\midrule
\multirow{5}{*}{\rotatebox[origin=c]{90}{\textit{DINOv1}}} & GCD & 74.1 & 89.8 & 66.3 & 52.5 & 72.5 & 42.2 \\
& SimGCD & 83.0 & 93.1 & 77.9 & 57.1 & 77.3 & 46.9 \\
& SPTNet & 85.4 & 93.2 & 81.4 & - & - & - \\
& SelEx & 83.1 & 93.6 & 77.8 & - & - & - \\
& Hyp-SelEx & 86.8 & 94.6 & 82.8 & - & - & - \\
\hline
\multirow{5}{*}{\rotatebox[origin=c]{90}{\textit{CLIP}}} & SimGCD$^\dagger$ & 86.1 & 94.5 & 81.9 & 48.2 & 72.7 & 36.0 \\
& CLIP-GCD & 84.0 & 95.5 & 78.2 & - & - & - \\
& TextGCD & 88.0 & 92.4 & 85.2 & 64.8 & \textbf{77.8} & 58.3 \\
& GET & 91.7 & 95.7 & 89.7 & 62.4 & 74.0 & 56.6 \\
& \textbf{Ours} & \textbf{92.1} & \textbf{96.0} & \textbf{90.2} & \textbf{66.7} & 77.3 & \textbf{61.1} \\
\bottomrule
\end{tabular}
}
\vskip -4pt
\end{table}

\begin{table*}[t]
\centering
\small
\caption{The average ACC (\%) on generic and fine-grained datasets. ``$^\dagger$'' denotes the reproduction of using CLIP backbone.}
\vskip -4pt
\label{tab:sota}
\setlength{\tabcolsep}{1.3mm}{
\begin{tabular}{ll ccc ccc ccc ccc ccc ccc ccc}
\toprule
& & \multicolumn{3}{c}{CIFAR-10} & \multicolumn{3}{c}{CIFAR-100} & \multicolumn{3}{c}{CUB} & \multicolumn{3}{c}{Stanford Cars} & \multicolumn{3}{c}{Oxford Pets} & \multicolumn{3}{c}{Flowers102} \\
\cmidrule(lr){3-5} \cmidrule(lr){6-8} \cmidrule(lr){9-11} \cmidrule(lr){12-14} \cmidrule(lr){15-17} \cmidrule(lr){18-20}
\multicolumn{2}{c}{Method} & All & Old & New & All & Old & New & All & Old & New & All & Old & New & All & Old & New & All & Old & New \\
\midrule
\multirow{7}{*}{\rotatebox[origin=c]{90}{\textit{DINOv1}}} & GCD & 91.5 & 97.9 & 88.2 & 73.0 & 76.2 & 66.5 & 51.3 & 56.6 & 48.7 & 39.0 & 57.6 & 29.9 & 80.2 & 85.1 & 77.6 & 74.4 & 74.9 & 74.1 \\
& GPC & 92.2 & 98.2 & 89.1 & 77.9 & 85.0 & 63.0 & 55.4 & 58.2 & 53.1 & 42.8 & 59.2 & 32.8 & - & - & - & - & - & - \\
& SimGCD & 97.1 & 95.1 & 98.1 & 80.1 & 81.2 & 77.8 & 60.3 & 65.6 & 57.7 & 53.8 & 71.9 & 45.0 & 87.7 & 85.9 & 88.6 & 71.3 & 80.9 & 66.5 \\
& PromptCAL & 97.9 & 96.6 & 98.5 & 81.2 & 84.2 & 75.3 & 62.9 & 64.4 & 62.1 & 50.2 & 70.1 & 40.6 & - & - & - & - & - & - \\
& SPTNet & 97.3 & 95.0 & \textbf{98.6} & 81.3 & 84.3 & 75.6 & 65.8 & 68.8 & 65.1 & 59.0 & 79.2 & 49.3 & - & - & - & - & - & - \\
& SelEx & 95.9 & 98.1 & 94.8 & 82.3 & 85.3 & 76.3 & 73.6 & 75.3 & 72.8 & 58.5 & 75.6 & 50.3 & 92.5 & 91.9 & 92.8 & - & - & - \\
& Hyp-SelEx & 96.7 & 97.6 & 96.3 & 82.4 & 85.1 & 77.0 & \textbf{79.8} & 75.8 & \textbf{81.8} & 62.9 & 80.0 & 54.7 & - & - & -  & - & - & - \\
\hline
\multirow{5}{*}{\rotatebox[origin=c]{90}{\textit{CLIP}}} & SimGCD$^\dagger$ & 96.6 & 94.7 & 97.5 & 81.6 & 82.6 & 79.5 & 62.0 & 76.8 & 54.6 & 75.9 & 81.4 & 73.1 & 88.6 & 75.2 & 95.7 & 75.3 & 87.8 & 69.0 \\
& CLIP-GCD & 96.6 & 97.2 & 96.4 & 85.2 & 85.0 & 85.6 & - & - & - & 62.8 & 77.1 & 55.7 & 70.6 & 88.2 & 62.2 & 76.3 & 88.6 & 70.2 \\
& TextGCD & 98.2 & 98.0 & 98.6 & 85.7 & \textbf{86.3} & 84.6 & 76.6 & \textbf{80.6} & 74.7 & 86.1 & 91.8 & 83.9 & 93.7 & 93.2 & 94.0 & 87.2 & 90.7 & 85.4 \\
& GET & 97.2 & 94.6 & 98.5 & 82.1 & 85.5 & 75.5 & 77.0 & 78.1 & 76.4 & 78.5 & 86.8 & 74.5 & 91.1 & 89.7 & 92.4 & 85.5 & 90.8 & 81.3 \\
& \textbf{Ours} & \textbf{98.5} & \textbf{98.3} & \textbf{98.6} & \textbf{86.4} & 86.2 & \textbf{86.9} & 78.3 & 78.5 & 78.2 & \textbf{89.2} & \textbf{93.1} & \textbf{87.3} & \textbf{95.7} & \textbf{95.1} & \textbf{96.0} & \textbf{93.5} & \textbf{93.3} & \textbf{93.9} \\
\bottomrule
\end{tabular}
}
\vskip -4pt
\end{table*}

\subsection{Performance on Benchmark Datasets}
We compare the performance of our SSR$^2$-GCD to uni-modal GCD methods, including GCD~\citep{Vaze:CVPR22-GCD}, GPC~\citep{Zhao:ICCV23-GPC}, SimGCD~\citep{Wen:ICCV23-SimGCD}, PromptCAL~\citep{Zhang:CVPR2023-promptCAL}, SPTNet~\citep{Wang:ICLR2024-SPTNet}, SelEx~\citep{Rastegar:ECCV24-SelEX} and HypCD~\citep{Liu:CVPR25-HypCD} with the SelEx backbone (denoted by ``Hyp-SelEx''), and multi-modal GCD methods, including CLIP-GCD~\citep{Rabha:Arxiv23-clipgcd}, TextGCD~\citep{Zheng:ECCV24-textgcd} and GET~\citep{Wang:CVPR25-get}.
Since that GET did not provide the results on Oxford Pets and Flowers102, we reproduced the results with %its 
the open-source code.
Experimental results are listed in Tables~\ref{tab:sota_2} and~\ref{tab:sota}. As can be read, our SSR$^2$-GCD consistently outperforms all other multi-modal counterparts on all test datasets.
We can also observe that our SSR$^2$-GCD narrows the accuracy gap between ``Old'' and ``New'' categories.
HypCD achieves the highest accuracy on CUB when using SelEx as the backbone, in which the embedding space is changed from %the 
Euclidean to %the 
hyperbolic, and thus it is complementary to our SSR$^2$-GCD. %method.
In addition, our SSR$^2$-GCD %method 
excels notably on Stanford Cars and Flowers102, achieving accuracy of 89.2\% and 93.5\% on ``All'' categories, outperforming all other baselines by 3.1\% and 6.3\%, respectively.
Note that CLIP performs poorly in out-of-domain datasets including Flowers102, achieving an accuracy of 70.4\% in zero-shot classification~\citep{Radford:ICML2021-CLIP}, 
whereas our SSR$^2$-GCD %method 
effectively refines the representations generated by CLIP to the target domain of Flowers102 and yields satisfactory performance.

\subsection{Evaluation on Representation Learning}
\label{subsec:eval_rep}

% \myparagraph{Naive Merging of Inter- and Intra-Modal Losses can be Ill-Advised}
% \myparagraph{On the Sub-optimality of Naively Merging Inter- and Intra-Modal Losses}}
%\myparagraph{Inter-Modal Alignment vs. Intra-Modal Alignment}

% here Mar 25，2026

\myparagraph{Alignment: Inter-Modal vs. Intra-Modal}
To evaluate the effect of using inter-modal and intra-modal alignments in our framework, we keep the text generation and classification methods the same and report the performance of using different losses for representation learning, including 
inter-modal loss (\ie, $\mathcal{L}_\text{CLIP}$ in Eq.~\eqref{eq:clip_inter_contrast}), 
intra-modal losses (\ie, $\mathcal{L}_\text{con}$ in Eq.~(\ref{eq:con_rep_loss}) and our $\mathcal{L}_\text{SSR$^2$}$ in Eq.~(\ref{eq:our_ssr_rep_learning})) and their combinations (\eg, $\mathcal{L}_\text{CLIP}+\mathcal{L}_\text{con}$).
The experimental results are listed in Table~\ref{tab:inter_intra_combine_ablation}. %As can be read that, we find 
We found that encouraging inter-modal alignment alone provides merely limited performance gain when compared to using the intra-modal losses alone, since that the learning of classifiers is based on the intra-modal relationships within each modality.
Specifically, our framework achieves the highest accuracy on five benchmark datasets when using the proposed $\mathcal{L}_\text{SSR$^2$}$.
Our framework trained with supervised and unsupervised contrastive loss $\mathcal{L}_\text{con}$ is still a strong intra-modal learning baseline, achieving the highest accuracy on CIFAR-100.
Interestingly, combining $\mathcal{L}_\text{CLIP}$ with intra-modal loss $\mathcal{L}_\text{con}$ or $\mathcal{L}_\text{SSR$^2$}$ cannot significantly improve the clustering accuracy and may even degrade it.
%
%Specifically, 
Note that $\mathcal{L}_\text{con}$ outperforms $\mathcal{L}_\text{CLIP}+\mathcal{L}_\text{con}$ on four datasets, except for CUB and Oxford Pets; whereas $\mathcal{L}_\text{SSR$^2$}$ surpasses $\mathcal{L}_\text{CLIP}+\mathcal{L}_\text{SSR$^2$}$ on all datasets, 
suggesting that explicitly imposing inter-modal alignment could deteriorate the intra-modal alignment.
\begin{table*}[tbh]
\centering
\small
\caption{Evaluation of different representation learning methods. Average ACC (\%) on ``All'' categories is reported. ``N/A'': frozen CLIP.}
\vskip -4pt
\setlength{\tabcolsep}{2mm}{
    \begin{tabular}{ l l l | c c c c c c}
    \toprule
    Rep. Losses & Inter & Intra & CIFAR-10 & CIFAR-100 & CUB & Stanford Cars & Oxford Pets & Flowers102 \\
    \midrule
    N/A & $\times$ & $\times$ & 97.9 & 84.1  & 74.5 & 86.0 & 91.9 & 87.4 \\
    $\mathcal{L}_\text{CLIP}$ & $\checkmark$ & $\times$ & 98.3 & 86.0 & 76.7 & 87.0 & 94.1 & 89.7 \\
    $\mathcal{L}_\text{con}$ & $\times$ & $\checkmark$ & \underline{98.4} & \textbf{86.7} & 77.5 & 87.9 & 94.9  & 91.8 \\
    $\mathcal{L}_\text{SSR$^2$}$ & $\times$ & $\checkmark$ & \textbf{98.5} & \underline{86.4} & \textbf{78.3} & \bf{89.2} & \textbf{95.7} & \textbf{93.5} \\
    $\mathcal{L}_\text{CLIP}$+$\mathcal{L}_\text{con}$ & $\checkmark$ & $\checkmark$ & 98.2 & 86.3 & \underline{78.0} & 86.7 & 95.0 & 90.9 \\
    $\mathcal{L}_\text{CLIP}$+$\mathcal{L}_\text{SSR$^2$}$ & $\checkmark$ & $\checkmark$ & 98.3 & 86.1 & 77.2 & \underline{88.1} & \underline{95.0} & \underline{92.9} \\

    \bottomrule
    \end{tabular}
}
\vskip -4pt
\label{tab:inter_intra_combine_ablation}
\end{table*}

\begin{figure}[th]
    \centering
    \includegraphics[height=1.6in]{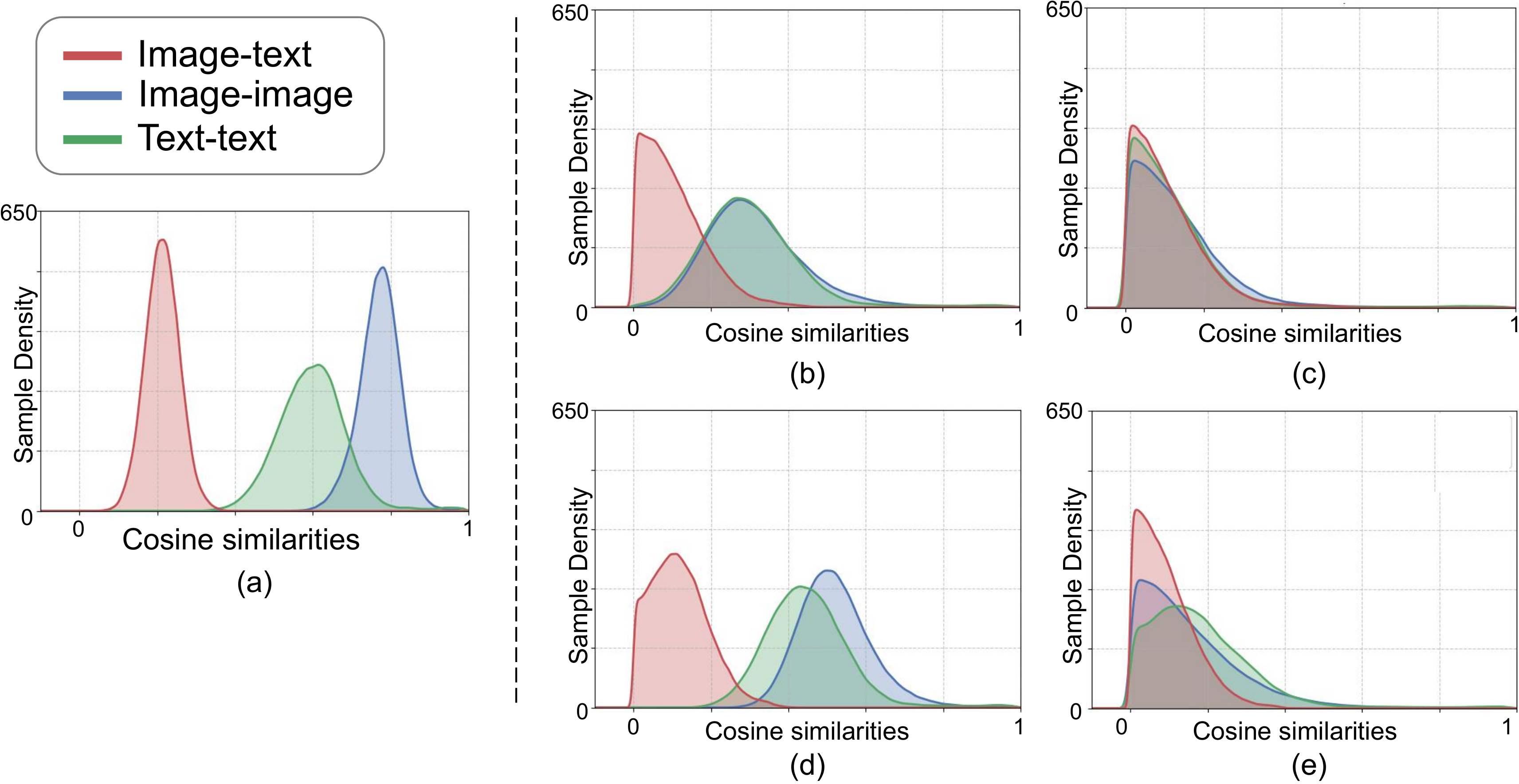}
    \caption{Distributions of pairwise similarities on Flowers102. (a): at the beginning of training. (b)-(c): training with $\mathcal{L}_\mathrm{SSR^2}$ at epochs 10 and 200. (d)-(e): training with $\mathcal{L}_\text{CLIP}$ at epochs 10 and 200.}
    \label{fig:density}
    \vskip -4pt
\end{figure}

\myparagraph{SSR$^2$ Works Well for Inter-modal Alignment and Intra-modal Alignment} %in multi-modal representation learning}
To %further 
evaluate the role of %proposed 
the SSR$^2$ objective in inter-modal and intra-modal alignments, we conduct a set of experiments to report the distributions of the pairwise similarity calculated %computed 
from image-text, image-image, and text-text pairs, %obtained from our framework, 
when using the loss $\mathcal{L}_\mathrm{SSR^2}$ or $\mathcal{L}_\text{CLIP}$ at the beginning of the training (\ie, epoch 0), at the end of the warm-up (\ie, epoch 10) and at the end of the training (\ie, epoch 200), respectively. 
Experimental results are shown in Figure~\ref{fig:density}. Clearly, we observe that the distributions of inter-modal similarities and intra-modal similarities exhibit substantial divergence at the beginning, where %while 
the distributions of image-image and text-text similarities within each modality also differ notably.
When using the proposed loss $\mathcal{L}_\mathrm{SSR^2}$, the distributions of intra-modal similarities are well aligned in the warm-up stage, and all distributions gaps almost vanish at the end of the training.
Specifically, optimizing the intra-modal alignment loss (\ie, $\mathcal{L}_\mathrm{SSR^2}$) alone still helps to align the distributions of image-text similarities. 
%
%By contrast, 
When using %the loss 
$\mathcal{L}_\text{CLIP}$, although the distributions of image-text similarities are aligned well, %whereas 
the image-image and text-text similarities, which are critical for GCD, are poorly aligned.

\begin{figure*}[t]
    \centering
    \begin{subfigure}[b]{0.245\linewidth}
           \includegraphics[width=\textwidth]{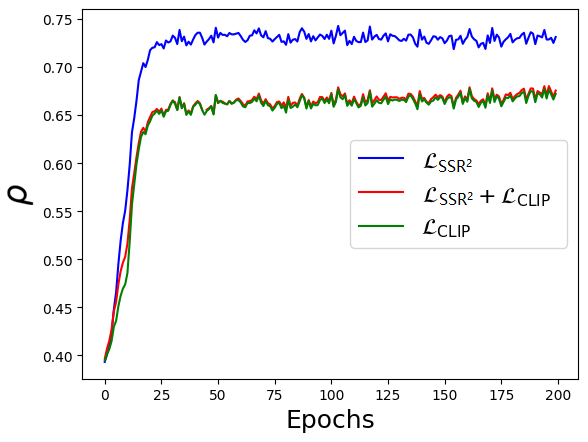}
            \caption{Visual modality}
    \end{subfigure}
    \begin{subfigure}[b]{0.245\linewidth}
           \includegraphics[width=\textwidth]{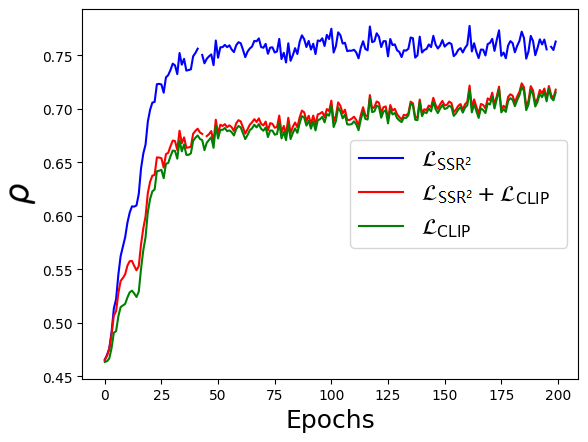}
            \caption{Textual modality}
    \end{subfigure}
    \begin{subfigure}[b]{0.245\linewidth}
           \includegraphics[width=\textwidth]{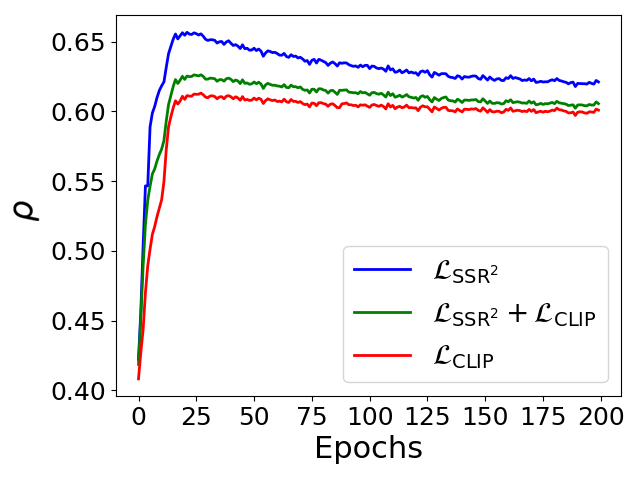}
            \caption{Image embeddings}
    \end{subfigure}
    \begin{subfigure}[b]{0.245\linewidth}
           \includegraphics[width=\textwidth]{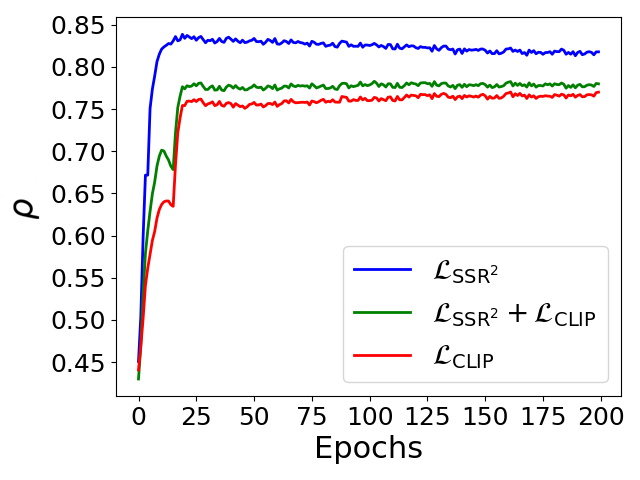}
            \caption{Text embeddings}
    \end{subfigure}
\caption{Consistency measure $\rho$ as a function of training epoch %Ratio curves 
when trained with different losses on different datasets. (a)-(b): on Flowers102. (c)-(d): on Stanford Cars.}
\label{Fig:sucr_ratio_of_edges}
\vskip -8pt
\end{figure*}

% \myparagraph{Evaluation on Intra-modal Alignment Behaviors during Training with Different Losses} 
\myparagraph{Evaluation on Intra-modal Alignment Behaviors}
For the reason why the intra-modal alignment loss works well in multi-modal representation learning, we account it in the implicit inter-modal alignment and the implicit hypothesis of clustering algorithms.
On one hand, the initial textual and visual representations are well aligned by using a pretrained CLIP backbone %model 
to select similar pseudo text-image pairs, and thus further encouraging inter-modal alignment between the embeddings of images and pseudo texts could be detrimental. %noisy.
%
% Furthermore, 
By contrast, minimizing %optimizing 
the proposed $\mathcal{L}_\mathrm{SSR^2}$ objective does not compromise the inter-modal alignment; %on the contrary, 
as shown in Figure~\ref{fig:density}, it implicitly promotes such an alignment. % through intra-modal mechanisms.
On the other hand, most GCD frameworks, including our SSR$^2$-GCD, %utilize 
employs the self-distillation loss for clustering. 
As shown %detailed 
in Eq.~(\ref{eq:detailed_simgcd_classifier}), the self-distillation loss %, which 
minimizes the cross-entropy between the predictions of the augmented data pairs from the same modality, %only involves 
involving only the consistency of the intra-modal embedding pairs.
In other words, %this loss 
using the loss in Eq.~(\ref{eq:detailed_simgcd_classifier}) is built on the assumption that the intra-modal embeddings of augmented data pairs should be close enough to be assigned to the same cluster. 
To quantify the intra-modal consistency, we construct the adjacency matrices $\mathbf W$ for each modality where %in which the weights on edges are the cosine similarities of the intra-modal embedding pairs, \ie, 
$W_{i,j}=\z^\top_i\z_j$.
Given the ground-truth categories $\{\C_1,\dots,\C_k\}$ indicated by their labels, one can quantify the consistency by computing the ratio of the weights on edges within each category to the weights on edges between different categories, \ie,
\begin{equation}
\label{eq:compactness-vs-cut}
    \rho=\frac{1}{k}\sum_{\ell=1}^k \rho_\ell, \quad 
    %\rho=\frac{1}{k}\sum_{\ell=1}^k \rho(\C_\ell), \quad 
%\rho(\C_\ell)
\rho_\ell=\frac{\sum_{i\in\C_\ell}\sum_{j\in\C_\ell}W_{i,j}}{\sum_{i\in\C_\ell}\sum_{j\notin\C_\ell}W_{i,j}}.
\end{equation}
Apparently, a larger value of $\rho$ indicates a higher ratio of correct intra-category links to incorrect inter-category links, reflecting better consistency of intra-modal embeddings.
%
%
% In Figure~\ref{Fig:sucr_ratio_of_edges} we show the $\rho$ curves with different representation losses (\ie, $\mathcal{L}_\text{CLIP}$, $\mathcal{L}_\text{SSR$^2$}$ and $\mathcal{L}_\text{CLIP}+\mathcal{L}_\text{SSR$^2$}$) on Flowers102.
We calculate the ratio %value 
$\rho$ at different epochs during the training period when using different losses %(\ie, $\mathcal{L}_\text{CLIP}$, $\mathcal{L}_\text{SSR$^2$}$ and $\mathcal{L}_\text{CLIP}+\mathcal{L}_\text{SSR$^2$}$) 
on Flowers102, and show it as a function of training epoch %and show the curves 
in Figure~\ref{Fig:sucr_ratio_of_edges}.
As can be seen, using $\mathcal{L}_\text{SSR$^2$}$ achieves higher $\rho$ than using $\mathcal{L}_\text{CLIP}$.
While $\mathcal{L}_\text{CLIP}$ can implicitly enhance intra-class discriminability to some extent, it is insufficient to ensure intra-modal alignment and thus yields %achieves less satisfactory 
inferior clustering accuracy.
This suggests %verifies 
that inter-modal alignment without any intra-modal constraint leads to intra-modal misalignment.
Moreover, we observe that the ratio curve of using $\mathcal{L}_\text{CLIP}+\mathcal{L}_\text{SSR$^2$}$ rises in parallel with that of using $\mathcal{L}_\text{SSR$^2$}$ in the early training stages, %of training, 
but eventually converges to %toward 
the curve of using $\mathcal{L}_\text{CLIP}$.
This verifies that the inter-modal alignment may damage the learning of intra-modal consistency.

\begin{figure}[t]
\vskip -6pt
    \centering
    \begin{subfigure}[b]{0.45\linewidth}
        \includegraphics[width=\textwidth]{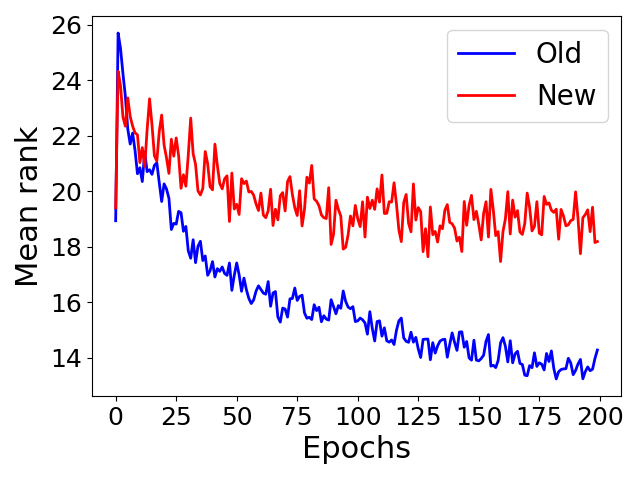}
        \caption{via $\mathcal{L}_\text{con}$ on Oxford Pets}
    \end{subfigure}
    \begin{subfigure}[b]{0.45\linewidth}
        \includegraphics[width=\textwidth]{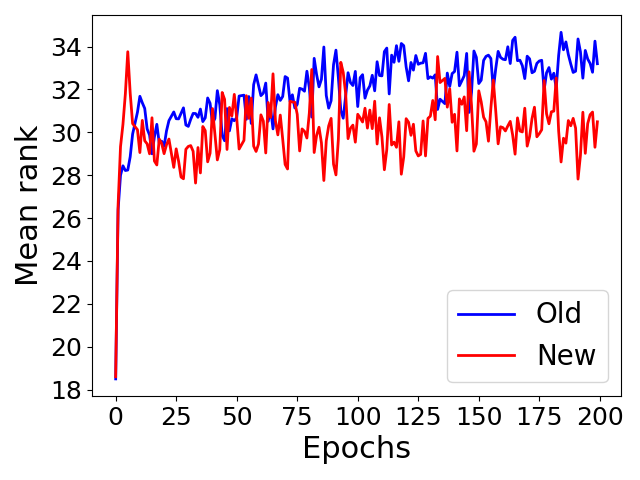}
        \caption{via $\mathcal{L}_\text{SSR$^2$}$ on Flowers102}
    \end{subfigure}
    \begin{subfigure}[b]{0.45\linewidth}
        \includegraphics[width=\textwidth]{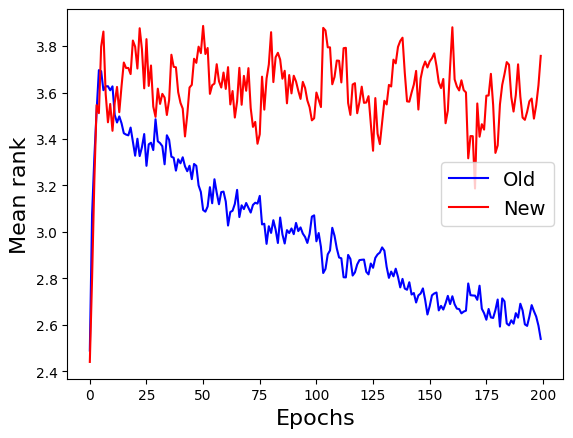}
        \caption{via $\mathcal{L}_\text{con}$ on Oxford Pets}
    \end{subfigure}
    \begin{subfigure}[b]{0.45\linewidth}
        \includegraphics[width=\textwidth]{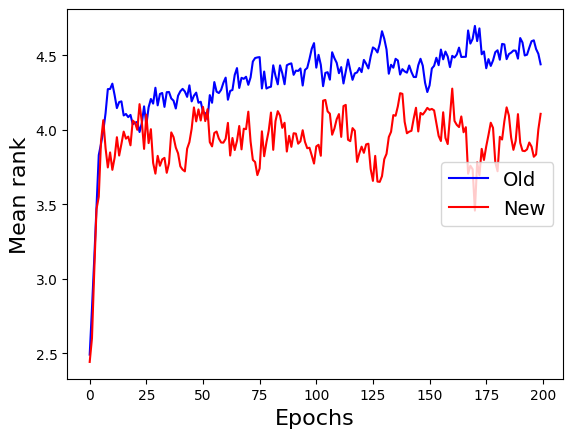}
        \caption{via $\mathcal{L}_\text{SSR$^2$}$ on Flowers102}
    \end{subfigure}
\caption{Mean ranks of image embeddings. % when trained via different losses on different datasets. 
%(a)-(b): on Oxford Pets. (c)-(d): on Flowers102.
}
\label{fig:rank_pets_flower}
\vskip -8pt
\end{figure}

\begin{figure}[t]
    \centering
    \begin{subfigure}[b]{0.4\linewidth}
        \includegraphics[width=0.99\textwidth]{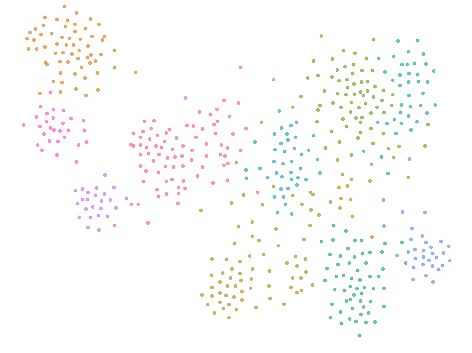}
        \caption{Frozen CLIP}
    \end{subfigure}\quad
    \begin{subfigure}[b]{0.4\linewidth}
        \includegraphics[width=0.99\textwidth]{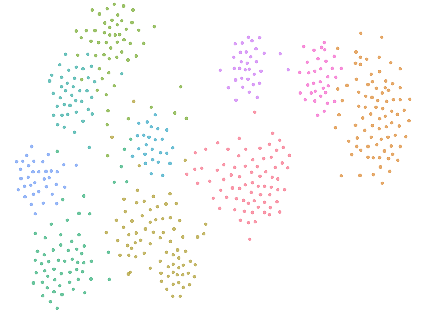}
        \caption{$\mathcal{L}_\text{CLIP}$}
    \end{subfigure} \\
    \begin{subfigure}[b]{0.4\linewidth}
        \includegraphics[width=0.99\textwidth]{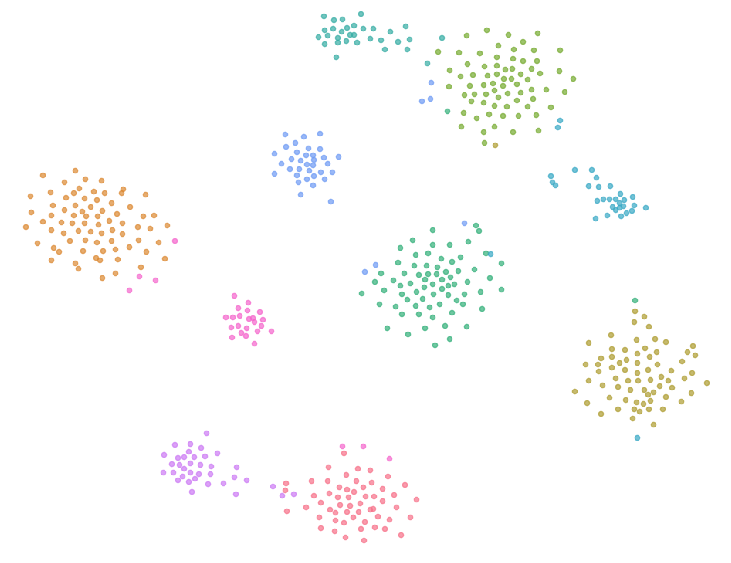}
        \caption{$\mathcal{L}_\text{con}$}
    \end{subfigure}\quad
    \begin{subfigure}[b]{0.4\linewidth}
        \includegraphics[width=0.99\textwidth]{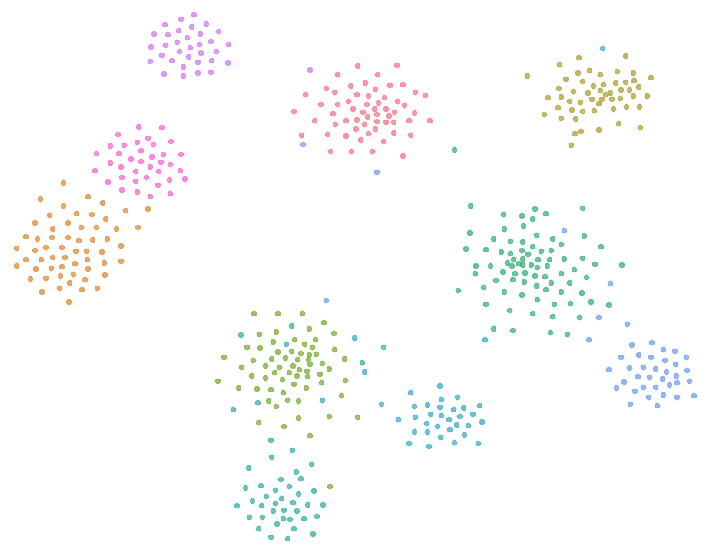}
        \caption{$\mathcal{L}_\text{SSR$^2$}$}
    \end{subfigure}
\caption{Visualizations of image embeddings.} %via different representation learning methods (\ie, using different losses).}
\label{fig:tsne}
\vskip -4pt
\end{figure}

\begin{figure}[h]
\vskip -2pt
\centering
\includegraphics[width=0.9\linewidth]{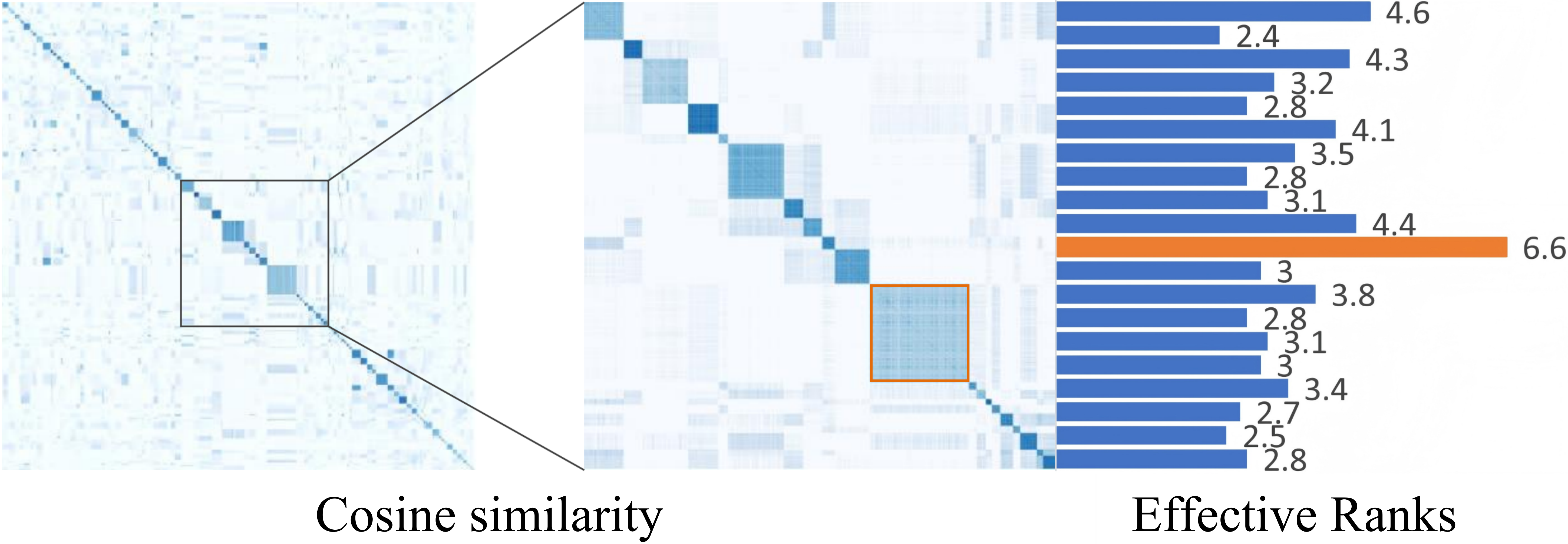}
\caption {Cosine similarity (ordered by ground-truths) and cluster ranks on Flowers102.}
\label{fig:cos_sim}
\vskip -12pt
\end{figure}

%\myparagraph{SSR$^2$ is a Compressor for Balanced Representation}}
\myparagraph{SSR$^2$ as a Compressor for Balanced Representation}
To evaluate the effectiveness of our %proposed 
$\mathcal{L}_\text{SSR$^2$}$, we conduct experiments to compare it to the widely used contrastive loss $\mathcal{L}_\text{con}$. 
To quantify the uniformity of the compression, we compute the numerical rank, % effective rank, 
%\footnote{Effective rank 
which is defined as the number of leading singular values whose cumulative energy proportion reaches 95\% of all singular values, of image embeddings per category,
\ie, $\{\texttt{rank}(\Z^\text{(i)}) \}^k_{i=1}$, where $\Z^\text{(i)}$ is the sub-matrix formed by the embeddings from the $i$-th ground-truth category.
We train our framework via using $\mathcal{L}_\text{con}$ and $\mathcal{L}_\text{SSR$^2$}$, respectively, %and plot 
compute the averaged ranks of ``Old'' categories and ``New'' categories, \ie, $\frac{1}{|\Y_L|}\sum_{i\in\Y_L}\texttt{rank}(\Z^{(i)})$ and $\frac{1}{|(\Y_U \setminus \Y_L)|}\sum_{j\in(\Y_U \setminus \Y_L)}\texttt{rank}(\Z^{(j)})$, on Flowers102 and Oxford Pets, %respectively, 
and display the results in Figure~\ref{fig:rank_pets_flower}.
We can see that the average rank of the ``Old'' categories dramatically decreases during the training period when using the loss $\mathcal{L}_\text{con}$, which is much lower than that of the ``New'' categories.
Intuitively, using the loss $\mathcal{L}_\text{con}$ will overly compress the embeddings toward the contrastive prototypes of the known categories. % and thereby damages the accuracy of clustering those data points in the boundaries of category manifolds.
In contrast, when training with the loss $\mathcal{L}_\text{SSR$^2$}$, the average ranks of the `Old'' and the ``New'' categories are preserved well, avoiding collapse.
%
%This indicates that the embeddings of each category are uniformly distributed, resulting in comparable clustering accuracy of `Old'' and ``New'' categories (see, e.g., Table~\ref{tab:sota}).
%
To have a better intuition, we use $t$-SNE~\citep{Van:JMLR08-tsne} to visualize the image embeddings from a frozen CLIP encoder, and trained via $\mathcal{L}_\text{CLIP}$, $\mathcal{L}_\text{con}$ and our $\mathcal{L}_\text{SSR$^2$}$ on balanced dataset CIFAR-10. Experimental results are shown in Figure~\ref{fig:tsne}.
%produced on CIFAR-10 by using a frozen CLIP encoder, by optimizing $\mathcal{L}_\text{CLIP}$, by optimizing $\mathcal{L}_\text{con}$ and by our approach. We display the experimental results in Figure~\ref{fig:tsne}.
%
% Clearly, we observe that our approach learns well-aligned and well-balanced representations that are evenly compressed within categories and discriminative between categories.
We observe that our approach learns well-aligned representations that are uniformly distributed across categories and discriminative between categories.
%
% \myparagraph{Performance of SSR$^2$-GCD on Imbalanced Categories}
To evaluate the performance of SSR$^2$-GCD on imbalanced datasets, we train our SSR$^2$-GCD on \textit{entire} Flowers102 (class-imbalanced) dataset and visualize the cosine similarity (ordered by ground-truths) and \textit{effective ranks} of the learned features in Figure~\ref{fig:cos_sim}.
As can be observed that, larger blocks (categories) tend to have higher ranks. This is because, the scaled compression term in the SSR$^2$ objectives tends to learn ``balanced'' embeddings, in which bigger categories span higher ranks with smaller singular values, smaller classes span lower ranks with larger singular values and the nonzero singular values within each class tend to be equal amplitude.
% with the scaled compression term in the SSR$^2$ objectives, our SSR$^2$-GCD tends to learn ``balanced'' embeddings, in which bigger categories span higher ranks with smaller singular values, smaller classes span lower ranks with larger singular values (and the nonzero singular values within each class tend to equal amplitude).
%

\subsection{Ablation Study}

To evaluate the effectiveness of each component in our SSR$^2$-GCD, %the proposed approach, 
we conduct a set of experiments on Stanford Cars and Flowers-102, in which the baseline is set up by using the most similar tag as pseudo-text, leveraging the frozen CLIP to generate the embeddings, and clustering both image and text embeddings via a single classifier with shared parameters.
Experimental results are listed in Table~\ref{tab:ablation_component}. We can read that the co-taught dual-branch classifiers outperform the baseline by identifying disparities between modalities.
The proposed RTA integrates richer information from tag and attribute candidates, thereby enhancing clustering performance on ``New'' categories. 
The intra-modal representation learning by using the loss $\mathcal{L}_\text{SSR$^2$}$ is also critical, since the training of classifiers relies on well-aligned intra-modal consisitency.
Our SSR$^2$-GCD achieves the best performance by integrating all the three components.

\begin{table}[h]
\small
\centering
\caption{Ablation study of different components.}
\vskip -8pt
\label{tab:ablation_component}
\setlength{\tabcolsep}{1.2mm}{
\begin{tabular}{ccccccccc}
\toprule
\multirow{2}{*}{Dual} & \multirow{2}{*}{RTA} & \multirow{2}{*}{$\mathcal{L}_\text{SSR$^2$}$} & \multicolumn{3}{c}{Stanford Cars} & \multicolumn{3}{c}{Flowers102} \\
\cmidrule(lr){4-6} \cmidrule(lr){7-9}
& & & All & Old & New & All & Old & New \\
\midrule
 $\times$ & $\times$ & $\times$ & 75.2 & 85.4 & 71.8 & 78.3 & 88.1 & 72.5 \\
$\checkmark$ & $\times$ & $\times$ & 81.7 & 90.3 & 77.1 & 83.9 & 88.3 & 81.3 \\
$\checkmark$ & $\checkmark$ & $\times$ & 86.0 & 91.0 & 83.4 & 87.4 & 90.8 & 85.5\\
$\checkmark$ & $\times$ & $\checkmark$ & 85.5 & 91.7 & 82.2 & 89.1 & 91.8 & 88.0 \\
$\checkmark$ & $\checkmark$ & $\checkmark$ & \textbf{89.2} & \textbf{93.1} & \textbf{87.3} & \textbf{93.5} & \textbf{93.3} & \textbf{93.9} \\
\bottomrule
\end{tabular}
}
\vskip -8pt
\end{table}

\subsection{More Evaluations}

\begin{figure}[h]
\vskip -6pt
    \centering
    \begin{subfigure}[b]{0.45\linewidth}
           \includegraphics[width=\textwidth]{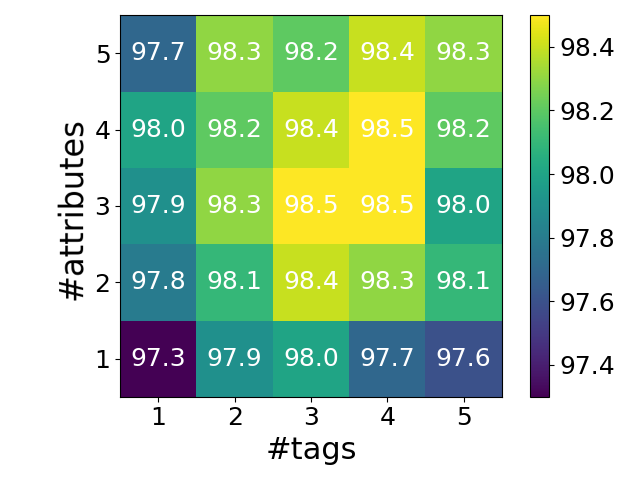}
            \caption{CIFAR-10}
    \end{subfigure}\quad
    \begin{subfigure}[b]{0.45\linewidth}
           \includegraphics[width=\textwidth]{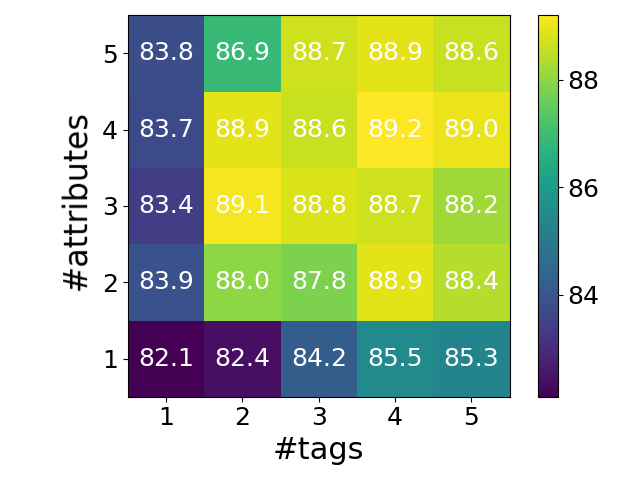}
            \caption{Stanford Cars}
    \end{subfigure}
\caption{\text{Effect of the number of candidates in RTA strategy.}}
\vskip -12pt
\label{Fig:rta_candidate}
\end{figure}

\myparagraph{Evaluation on RTA} %Retrieval-based Text Aggregation (RTA)}
In our proposed RTA, %Retrieval-based Text Aggregation (RTA), 
we introduce a hyper-parameter $\alpha$ %serves as a balance factor of integrating 
to balance the use of the most-similar candidates and other candidates.
To evaluate the sensitivity to %hyper-parameter 
$\alpha$, we report the accuracy of ``All'' categories on four generic and fine-grained datasets, with varying values of $\alpha\in \{0.2,0.3,0.4,0.5,0.6,0.7,0.8\}$ while keeping other components in our SSR$^2$-GCD unchanged.
As shown in Table~\ref{tab:rta_parameter}, we observe that using the proposed RTA achieves the best performance on CIFAR-10, Stanford Cars and Flowers102 with $\alpha=0.5$, and achieves the best performance on CIFAR-100 with $\alpha=0.4$.
\begin{table}[]
\small
\vskip -10pt
    \centering
    \caption{Effect of the hyper-parameter $\alpha$ in RTA strategy.}
    \vskip -6pt
    \setlength{\tabcolsep}{1.5mm}{
    \begin{tabular}{l|cccccccc}
        \toprule
        \diagbox[height=1.7em]{Data}{$\alpha$} & 0.2 & 0.3 & 0.4 & 0.5 & 0.6 & 0.7 & 0.8  \\
        \midrule
        CIFAR-10 & 98.3 & 98.4 & 98.4 & \textbf{98.5} & 98.3 & 98.2 & 97.8 \\
        CIFAR-100 & 86.0 & 86.6 & \textbf{86.7} & 86.4 & 85.8 & 85.2 & 84.1 &  \\
        Stanford Cars & 87.1 & 88.7 & 88.9 & \textbf{89.2} & 88.8 & 87.2 & 86.4 \\
        Flowers102 & 92.9 & 93.3 & 93.3 & \textbf{93.5} & 92.8 & 92.1 & 91.5\\
        \bottomrule
    \end{tabular}
    }
    \label{tab:rta_parameter}
    \vskip -12pt
\end{table}
Moreover, in RTA as shown in Eq.(\ref{eq:RTA}), we select multiple candidates for tags and attributes. 
To evaluate the effect of the number of candidates, we report the accuracy of ``All'' categories on CIFAR-10 and Stanford Cars, with varying number of tags and attributes in $\{1,2,3,4,5\}$, respectively. 
% while keeping other components in our SSR$^2$-GCD unchanged.
%
As shown in Figure~\ref{Fig:rta_candidate}, we can read that integrating more %information from the 
candidates yields consistent performance improvements.
Specifically, %the proposed RTA 
the performance achieves the best %performance 
when the number of tags and attributes is set to 3 or 4, respectively. More evaluations are provided in the Supporting Materials.

% \begin{figure}[h]
%     \centering
%     \begin{subfigure}[b]{0.5\linewidth}
%            \includegraphics[width=\textwidth]{figs/cifar10_candidate_heatmap.png}
%             \caption{CIFAR-10}
%     \end{subfigure}\hfill
%     \begin{subfigure}[b]{0.5\linewidth}
%            \includegraphics[width=\textwidth]{figs/cifar100_candidate_heatmap.png}
%             \caption{CIFAR-100}
%     \end{subfigure}\hfill
%     \begin{subfigure}[b]{0.5\linewidth}
%            \includegraphics[width=\textwidth]{figs/scars_candidate_heatmap.png}
%             \caption{Stanford Cars}
%     \end{subfigure}\hfill
%     \begin{subfigure}[b]{0.5\linewidth}
%            \includegraphics[width=\textwidth]{figs/flowers_candidate_heatmap.png}
%             \caption{Flowers102}
%     \end{subfigure}
% \caption{\text{Effect of the number of candidates in RTA strategy.}}
% \label{Fig:rta_candidate}
% \end{figure}

\section{Conclusion}
\label{sec:conclusion}
We proposed a novel and effective multi-modal representation learning framework, called SSR$^2$-GCD. %to tackle with the multi-modal GCD task. In particular, 
To be specific, we incorporated a semi-supervised rate reduction loss to learn balanced structural representations for multi-modal GCD task. % that are evenly compressed across categories.
Moreover, we designed a retrieval-based text aggregation strategy to enhance text generation and thus boost knowledge transfer.
We conducted extensive experiments on eight benchmark datasets, and experimental results demonstrated the effectiveness of our proposed SSR$^2$-GCD approach and verified that prioritizing intra-modal consistency via SSR$^2$ is beneficial for multi-modal GCD.

%\

%\myparagraph{Limitation}
% We note that the time and memory burden of our approach will slightly increase when the number of candidates is enlarged due to using multiple tag and attribute candidates as the input to CLIP encoders. 
%
%Besides, 
%In our framework, the image and text cues are treated equally. Thus a further exploration on the importance of each modality in the multi-modal GCD task is left for future work.

%\
\section*{Acknowledgments}
%\myparagraph{Acknowledgments}
This work is supported by the National Natural Science Foundation of China under Grant No.~62576048.
%Chun-Guang Li is the corresponding author. % 61876022 

{
\small
\bibliographystyle{ieeenat_fullname}
\bibliography{references}
}

% Begin: appendix
\clearpage
\appendix

% Reset numbers to zero
\setcounter{table}{0}
\setcounter{figure}{0}

% A.1, B.1 style
\renewcommand{\thetable}{\Alph{section}.\arabic{table}}
\renewcommand{\thefigure}{\Alph{section}.\arabic{figure}}

%}
%\title{Supporting Materials}
{\mcb 
\section{Experimental Details}

\subsection{Datasets}
In Table~\ref{tab:data}, we provide a statistical summarization of the eight generic and fine-grained datasets.
Among these benchmarks, the generic datasets including CIFAR-10, CIFAR-100, ImageNet-100 and ImageNet-1k consist of categories of open-world, whereas the fine-grained benchmarks including CUB, Stanford Cars, Oxford Pets, and Flowers102 are largely domain-specific.
Following the GCD protocol~\citep{Vaze:CVPR22-GCD}, we select the first half of classes as the known categories for each benchmark dataset, except for CIFAR-100, in which we select the first 80 classes as the known categories.

\begin{table}[ht]
    \centering
    \caption{Statistics of the benchmark datasets.}
    \label{tab:data}
    \begin{tabular}{l r r r r}
        \toprule
        \multirow{2}{*}{Dataset} & \multicolumn{2}{c}{Labelled} & \multicolumn{2}{c}{Unlabelled} \\
        \cmidrule(lr){2-3} \cmidrule(lr){4-5}
         & \#Image & \#Class & \#Image & \#Class \\
        \midrule
        CIFAR10 & 12.5K & 5 & 37.5K & 10 \\
        CIFAR100 & 20.0K & 80 & 30.0K & 100 \\
        ImageNet-100 & 31.9K & 50 & 95.3K & 100 \\
        ImageNet-1K & 321K & 500 & 960K & 1000 \\
        CUB & 1.5K & 100 & 4.5K & 200 \\
        Stanford Cars & 2.0K & 98 & 6.1K & 196 \\
        Oxford Pets & 1.9K & 19 & 5.5K & 37 \\
        Flowers102 & 0.3K & 51 & 0.8K & 102 \\
        
        \bottomrule
    \end{tabular}
\end{table}

\subsection{Experiments Settings}

\myparagraph{Network Architecture}
In Figure~\ref{fig:rta_arch}, we present a detailed overview of our proposed Retrieval-based Text Aggregation (RTA).
In Table~\ref{tab:arch}, we present the learnable parameters of network architecture. % in the CLIP encoders and classifiers.
The visual and textual branches share the same architecture.
Specifically, when using CLIP-B/16 as the backbone, we fine-tune the last residual attention block (which includes the multi-head self-attention mechanism, feed-forward network, and layer normalization), along with the image and text projectors of CLIP.
Additionally, the dual-branch classifiers are learned with the fine-tuning of CLIP jointly.

\begin{figure}[ht]
    \centering
    \includegraphics[width=0.9\linewidth]{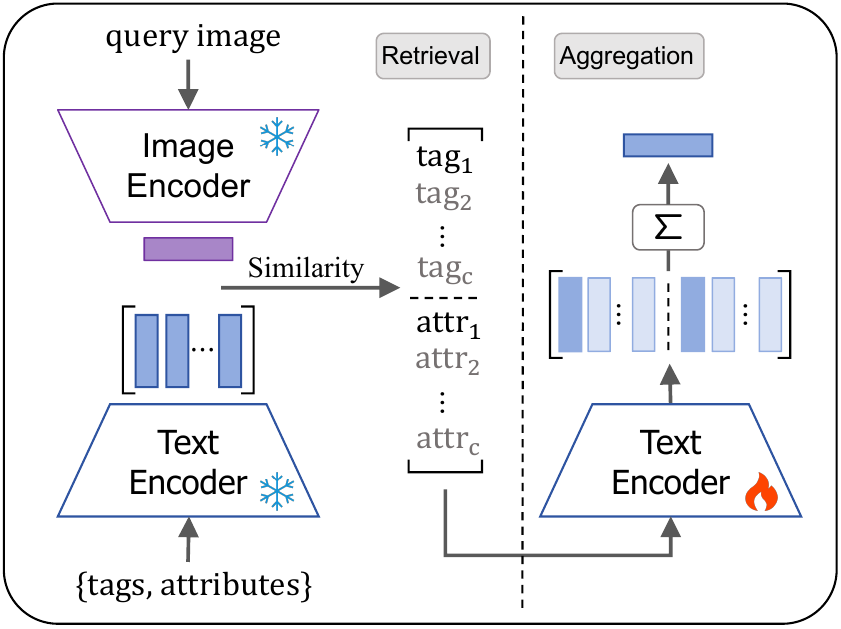}
    \caption{An overview of Retrieval-based Text Aggregation (RTA).}
    \label{fig:rta_arch}
\end{figure}

\begin{table}
\centering
\caption{Model architectures. ``\textit{CLIP}'' denotes the learnable parameters in CLIP and ``\textit{Cls}'' denotes the classifiers.}
\label{tab:arch}
\setlength{\tabcolsep}{1.5mm}{
    \begin{tabular}{l c}
            \toprule
            \multirow{2}{*}{\rotatebox[origin=c]{90}{\textit{CLIP}}} & Last residual attention block: $\RR^{512}\rightarrow \RR^{512}$ \\
            & Image/text projector: $\RR^{512}\rightarrow \RR^{512}$ \\
            \midrule
            \multirow{2}{*}{\rotatebox[origin=c]{90}{\textit{Cls.}}} & Linear projection: $\RR^{512}\rightarrow \RR^{K}$ \\
            & Softmax function \\
            \bottomrule
    \end{tabular}

}
\end{table}

\begin{table}[h]
    \centering
    \footnotesize
    \caption{Python code for the image augmentation. }
    \begin{tabular}{l}
        \toprule
        \texttt{from torchvision.transforms import *} \\
        \midrule
        \texttt{Compose([} \\
        \texttt{\quad RandomResizedCrop(32, BILINEAR)} \\
        \texttt{\quad RandomHorizontalFlip(p=0.5), } \\
        \texttt{\quad ColorJitter() } \\
        \texttt{\quad ToTensor(),} \\
        \texttt{\quad Normalize([0.485, 0.456, 0.406],} \\
        \texttt{\quad\quad\quad\quad\quad\quad\quad [0.229, 0.224, 0.225])} \\
        \quad \texttt{)])} \\
        \bottomrule
    \end{tabular}
    \label{tab:aug_img}
\end{table}

% Times New Roman
% \begin{table}[th]
%     \centering
%     \footnotesize
%     \caption{Pseudo-code for the text augmentation. }
%     \begin{tabular}{l}
%     \toprule
%     \textbf{Text Augmentation Strategy} \\
%     \midrule
%     \textbf{Input:} text \\
%     \textbf{For each} word \textbf{in} text: \\
%     \quad \textbf{If} len(word) $\geq$ 3: \\
%     \quad \quad \textit{index} $\gets$ random(1, len(word)$-$2) \\
%     \quad \quad \textit{action} $\gets$ random( \{replace, delete, add, none\} ) \\
%     \quad \quad \textbf{Case} \textit{action}: \\
%     \quad \quad \quad replace: word $\gets$ replace random char at \textit{index} \\
%     \quad \quad \quad delete: word $\gets$ remove char at \textit{index} \\
%     \quad \quad \quad add: word $\gets$ insert random char at \textit{index} \\
%     \quad \quad \quad none: \textbf{continue} \\
%     \textbf{Output:} augmented text \\
%     \bottomrule
%     \end{tabular}
%     \label{tab:aug_txt}
% \end{table}

% texttt
\begin{table}[th]
    \centering
    \footnotesize
    \caption{Pseudo-code for the text augmentation.}
    \begin{tabular}{l}
    \toprule
    %\textbf{\texttt{Text Augmentation Strategy}} \\
    \textbf{\texttt{Input:}} \texttt{text} \\
    \midrule
    \textbf{\texttt{For each}} \texttt{word} \textbf{\texttt{in}} \texttt{text}: \\
    \quad \textbf{\texttt{If}} \texttt{len(word)} $\geq$ \texttt{3}: \\
    \quad \quad \textit{\texttt{index}} $\gets$ \texttt{random(1, len(word)$-$2)} \\
    \quad \quad \textit{\texttt{action}} $\gets$ \texttt{random(\{replace, delete, add, none\})} \\
    \quad \quad \textbf{\texttt{Case}} \textit{\texttt{action}}: \\
    \quad \quad \quad \texttt{replace}: \texttt{word} $\gets$ \texttt{replace random char at} \textit{\texttt{index}} \\
    \quad \quad \quad \texttt{delete}: \texttt{word} $\gets$ \texttt{remove char at} \textit{\texttt{index}} \\
    \quad \quad \quad \texttt{add}: \texttt{word} $\gets$ \texttt{insert random char at} \textit{\texttt{index}} \\
    \quad \quad \quad \texttt{none}: \textbf{\texttt{continue}} \\
    \midrule
    \textbf{\texttt{Output:}} \texttt{augmented text} \\
    \bottomrule
    \end{tabular}
    \label{tab:aug_txt}
\end{table}

\begin{table*}[]
\centering
\small
\caption{The $\text{mean}\pm \text{std}$ ACC (\%) of TextGCD, GET and our approach on generic datasets.}
\label{tab:std_1}
\setlength{\tabcolsep}{1.mm}{
\begin{tabular}{ll ccc ccc ccc ccc ccc}
\toprule
& &  \multicolumn{3}{c}{CIFAR-10} & \multicolumn{3}{c}{CIFAR-100} & \multicolumn{3}{c}{ImageNet-100} & \multicolumn{3}{c}{ImageNet-1k} \\
\cmidrule(lr){3-5} \cmidrule(lr){6-8} \cmidrule(lr){9-11} \cmidrule(lr){12-14}
\multicolumn{2}{c}{Method} & All & Old & New & All & Old & New & All & Old & New & All & Old & New \\
\midrule

& TextGCD & 98.2{$\scriptstyle\pm 0.0$} & 98.0{$\scriptstyle\pm 0.2$} & \textbf{98.6}{$\scriptstyle\pm 0.1$} & 85.7{$\scriptstyle\pm 0.9$} & \textbf{86.3}{$\scriptstyle\pm 0.6$} & 84.6{$\scriptstyle\pm 2.2$} & 88.0{$\scriptstyle\pm 0.6$} & 92.4{$\scriptstyle\pm 0.9$} & 85.2{$\scriptstyle\pm 1.2$} & 64.8{$\scriptstyle\pm 0.2$} & \textbf{77.8}{$\scriptstyle\pm 0.5$} & 58.3{$\scriptstyle\pm 0.4$} \\
& GET & 97.2{$\scriptstyle\pm 0.1$} & 94.6{$\scriptstyle\pm 0.1$} & 98.5{$\scriptstyle\pm 0.1$} & 82.1{$\scriptstyle\pm 0.4$} & 85.5{$\scriptstyle\pm 0.5$} & 75.5{$\scriptstyle\pm 0.5$} & 91.7{$\scriptstyle\pm 0.3$} & 95.7{$\scriptstyle\pm 0.0$} & 89.7{$\scriptstyle\pm 0.4$} & 62.4{$\scriptstyle\pm 0.0$} & 74.0{$\scriptstyle\pm 0.2$} & 56.6{$\scriptstyle\pm 0.1$} \\
& \textbf{Ours} & \textbf{98.5}{$\scriptstyle\pm 0.2$} & \textbf{98.3}{$\scriptstyle\pm 0.2$} & \textbf{98.6}{$\scriptstyle\pm 0.3$} & \textbf{86.4}{$\scriptstyle\pm 0.5$} & 86.2{$\scriptstyle\pm 0.6$} & \textbf{86.9}{$\scriptstyle\pm 1.6$} & \textbf{92.1}{$\scriptstyle\pm 0.3$} & \textbf{96.0}{$\scriptstyle\pm 0.6$} & \textbf{90.2}{$\scriptstyle\pm 0.8$} & \textbf{66.7}{$\scriptstyle\pm 0.1$} & 77.3{$\scriptstyle\pm 0.3$} & \textbf{61.1}{$\scriptstyle\pm 0.2$} \\
\bottomrule
\end{tabular}
}
\end{table*}

\begin{table*}[th]
\centering
\small
\caption{The $\text{mean}\pm \text{std}$ ACC (\%) of TextGCD, GET and our approach on fine-grained datasets.}
\label{tab:std_2}
\setlength{\tabcolsep}{1.mm}{
\begin{tabular}{ll ccc ccc ccc ccc ccc}
\toprule
& &  \multicolumn{3}{c}{CUB} & \multicolumn{3}{c}{Stanford Cars} & \multicolumn{3}{c}{Oxford Pets} & \multicolumn{3}{c}{Flowers102} \\
\cmidrule(lr){3-5} \cmidrule(lr){6-8} \cmidrule(lr){9-11} \cmidrule(lr){12-14}
\multicolumn{2}{c}{Method} & All & Old & New & All & Old & New & All & Old & New & All & Old & New \\
\midrule

& TextGCD & 76.6{$\scriptstyle\pm 0.6$} & \textbf{80.6{$\scriptstyle\pm 2.0$}} & 74.7{$\scriptstyle\pm 1.7$} & 86.1{$\scriptstyle\pm 0.9$} & 91.8{$\scriptstyle\pm 0.4$} & 83.9{$\scriptstyle\pm 1.3$} & 93.7{$\scriptstyle\pm 0.6$} & 93.2{$\scriptstyle\pm 1.1$} & 94.0{$\scriptstyle\pm 0.9$} & 87.2{$\scriptstyle\pm 2.3$} & 90.7{$\scriptstyle\pm 1.3$} & 85.4{$\scriptstyle\pm 3.8$} \\
& GET & 77.0{$\scriptstyle\pm 0.5$} & 78.1{$\scriptstyle\pm 1.6$} & 76.4{$\scriptstyle\pm 1.2$} & 78.5{$\scriptstyle\pm 1.3$} & 86.8{$\scriptstyle\pm 1.5$} & 74.5{$\scriptstyle\pm 2.2$} & 91.1{$\scriptstyle\pm 1.0$} & 89.7{$\scriptstyle\pm 1.6$} & 92.4{$\scriptstyle\pm 1.2$} & 85.5{$\scriptstyle\pm 0.5$} & 90.8{$\scriptstyle\pm 1.5$} & 81.3{$\scriptstyle\pm 1.7$} \\
& \textbf{Ours} & \textbf{78.3{$\scriptstyle\pm 0.8$}} & 78.5{$\scriptstyle\pm 1.2$} & \textbf{78.2{$\scriptstyle\pm 0.9$}} & \textbf{89.2{$\scriptstyle\pm 0.3$}} & \textbf{93.1{$\scriptstyle\pm 0.9$}} & \textbf{87.3{$\scriptstyle\pm 0.2$}} & \textbf{95.7{$\scriptstyle\pm 0.2$}} & \textbf{95.1{$\scriptstyle\pm 0.5$}} & \textbf{96.0{$\scriptstyle\pm 0.4$}} & \textbf{93.5{$\scriptstyle\pm 0.4$}} & \textbf{93.3{$\scriptstyle\pm 0.8$}} & \textbf{93.9{$\scriptstyle\pm 1.0$}} \\
\bottomrule
\end{tabular}
}
\end{table*}

\myparagraph{Data Preparation}
For each mini-batch, %data, 
we generate a set of text embeddings for query images by integrating four similar tags and four  similar attributes through the proposed RTA strategy %retrieval-based text aggregation 
by using the encoders of CLIP-H/14 for all test datasets.
Then, both images and text are augmented into two views, and the augmentation strategies are the same across datasets, as detailed in Tables~\ref{tab:aug_img} and~\ref{tab:aug_txt}.
The embeddings of augmented images and text are used for representation learning.

\myparagraph{Training Settings}
We use %the the 
Stochastic Gradient Descent (SGD) with the momentum of $0.9$, the weight decay of $1\times 10^{-4}$, and using cosine annealing learning rate decay for the training process.
We train the model for $200$ epochs and set the batch size to $128$.
The random seeds for the three trials are set to $\{0,1,2\}$.
For representation learning, the initial learning rate for CLIP is set to $0.001$, and our proposed objective $\mathcal{L}_\mathrm{SSR^2}$ does not need %introduce 
an extra balancing hyper-parameter, except for $\epsilon$ which is set as 0.5.
%
% For clustering, the setting follows the default configurations as in SimGCD and TextGCD.
%
The initial learning rate for the two classifiers is set to $0.1$, the epochs for warm-up stage is set to $10$, and $60\%$ of pseudo-labels of each categories are selected for co-teaching.
We use the same setting for all test datasets. 
All experiments are conducted on a single NVIDIA GeForce RTX3090 GPU.

\section{More Experiments} % Results}
\subsection{Evaluation on Model Stability}
In Tables~\ref{tab:std_1} and~\ref{tab:std_2}, we report the mean accuracy with standard deviation ($\pm\text{std}$) %accuracies 
over 3 trials across the test datasets, and compare to the multi-modal counterparts TextGCD~\citep{Zheng:ECCV24-textgcd} and GET~\citep{Wang:CVPR25-get}.
As can be seen, the performance of our proposed SSR$^2$-GCD is relatively stable. % with a lower standard deviation.

\subsection{Effect of Retrieval-based Text Aggregation}

\begin{table}[h]
  \centering
  \small
  \caption{The number of identical tags between the candidate pool and the unknown categories in four benchmark datasets.}
  \label{tab:imagenet_matching}
  \begin{tabular}{lcc}
    \toprule
    Dataset & \# Unknown categories & \# Removed tags  \\
    \midrule
    CUB-200 & 100 & 90  \\
    Stanford Cars & 98 & 83  \\
    Flowers102 & 51 & 43  \\
    Oxford Pets & 19 & 17  \\
    \bottomrule
  \end{tabular}
\end{table}

In Table~\ref{tab:imagenet_matching}, we report the results of our SSR$^2$-GCD compared to TextGCD~\citep{Zheng:ECCV24-textgcd} when removing the prompts of the \textit{unknown} categories for each dataset from the candidate pool, since that the prompts in candidate pool may semantically identical to the unknown categories.
As can be seen in Table~\ref{tab:without}, though the performance of both methods is slightly dropped, our SSR$^2$-GCD is still leading. 
% For a fair comparison to other retrieval based methods, we use same candidate pool and backbones as in TextGCD to search prompt candidates in the main manuscript. 

\begin{table}[h]
\centering
\small
\caption{Effect of retrieval-based methods with or without (w/o) prompts from unknown categories on four test datasets. ACC (\%) on ``All'' categories is reported.}
\label{tab:without}
    \begin{tabular}{l|cccc}
        \hline
        Datasets & CUB & Cars & Pets & Flowers \\
        \hline
        TexGCD~\citep{Zheng:ECCV24-textgcd} & $76.6$ & $86.1$ & $93.7$ & $87.2$ \\
        TextGCD (w/o) & $75.7$ & $85.5$ & $91.5$ & $84.0$ \\
        % SSR
        SSR$^2$-GCD (w/o) & $\bf 77.6$ & $\bf 88.1$ & $\bf 94.0$ & $\bf 91.6$ \\
        \hline
    \end{tabular}
\end{table}

\subsection{Evaluation on Modality Alignments for Representation Learning}

\myparagraph{Evaluation on More Inter-Modal Alignment Methods}
To further evaluate the necessity of inter-modal alignment, we report the performance of using different inter-modal alignment %representation 
loss for training our SSR$^2$-GCD. % framework.
Specifically, we introduce a Cross-modal %inter-modal 
Instance Consistency Objective (CICO) which is proposed in GET~\citep{Wang:CVPR25-get} %to serve 
as an inter-modal alignment constraint. CICO is defined as follows:
\begin{equation}
    \label{eq:cico}
    \mathcal{L}_{\text{CICO}} = \frac{1}{2|B|} \sum_{i \in B} \left( D_{\text{KL}}(\s_i^\text{T} \| \s_i^\text{I}) + D_{\text{KL}}(\s_i^\text{I} \| \s_i^\text{T}) \right),
\end{equation}
where $D_{\text{KL}}$ is the Kullback-Leibler divergence, $B$ denotes the mini-batch data, $\s_i^\text{I} = \text{softmax}({\z_i^\text{I}}^\top \A^\text{I})$ and $\s_i^\text{T} = \text{softmax}({\z_i^\text{T}}^\top \A^\text{T})$ measure the similarity between the $i$-th image/text embeddings and prototypes, and $\mathcal{A}^\text{I}$ and $\mathcal{A}^\text{T}$ are the prototypes determined by %calculated using 
the labeled anchors for each modality. 

In Table~\ref{tab:inter_intra_combine_added}, we report the performance of using $\mathcal{L}_\text{CICO}$ and its combination with the intra-modal alignment losses for representation learning, in which the results are marked in gray. 
As can be seen that, both inter-modal alignment losses $\mathcal{L}_\text{CLIP}$ and $\mathcal{L}_\text{CICO}$ fail to bring %provide 
performance improvements when combining with the intra-modal alignment loss. This further confirms that %verifying the argument that 
performing extra inter-modal alignment is not necessary.

\begin{table}[h]
\centering
\small
\caption{Evaluation of different representation learning methods. Average ACC (\%) on ``All'' categories is reported. ``N/A'' denotes using a frozen CLIP.}
\label{tab:inter_intra_combine_added}
\setlength{\tabcolsep}{.7mm}{
    \begin{tabular}{ l l l | c c c c c c}
    \toprule
    Rep. Losses & Inter & Intra & C-10 & C-100 & CUB & Cars & Pets & Flowers \\
    \midrule
    N/A & $\times$ & $\times$ & 97.9 & 84.1  & 74.5 & 86.0 & 91.9 & 87.4 \\
    $\mathcal{L}_\text{CLIP}$ & $\checkmark$ & $\times$ & 98.3 & 86.0 & 76.6 & 86.7 & 93.9 & 89.7 \\
    
    \rowcolor{gray!20}
    $\mathcal{L}_\text{CICO}$ & $\checkmark$ & $\times$ & 98.0 & 85.0 & 76.4 & 86.1 & 94.9 & 87.2 \\
    \hline
    $\mathcal{L}_\text{con}$ & $\times$ & $\checkmark$ & \underline{98.4} & \textbf{86.7} & 77.5 & 87.9 & 94.9  & 91.8 \\
    $\mathcal{L}_\mathrm{SSR^2}$ & $\times$ & $\checkmark$ & \textbf{98.5} & \underline{86.4} & \textbf{78.3} & \bf{89.2} & \textbf{95.7} & \textbf{93.5} \\
    \hline
    $\mathcal{L}_\text{CLIP}$+$\mathcal{L}_\text{con}$ & $\checkmark$ & $\checkmark$ & 98.2 & 86.3 & \underline{78.0} & 86.7 & 95.0 & 90.9 \\
    \rowcolor{gray!20}
    $\mathcal{L}_\text{CICO}$+$\mathcal{L}_\text{con}$ & $\checkmark$ & $\checkmark$ & \underline{98.4} & 85.9 & 76.8 & 87.0 & 94.4 & 88.6 \\
    $\mathcal{L}_\text{CLIP}$+$\mathcal{L}_\mathrm{SSR^2}$ & $\checkmark$ & $\checkmark$ & 98.3 & 86.1 & 77.2 & \underline{88.1} & 95.0 & \underline{92.9} \\
    \rowcolor{gray!20}
    $\mathcal{L}_\text{CICO}$+$\mathcal{L}_\mathrm{SSR^2}$ & $\checkmark$ & $\checkmark$ & 98.3 & 86.1 & 76.7 & 87.5  & \underline{95.5} & 92.1 \\

    \bottomrule
    \end{tabular}
}
\end{table}

\begin{table*}[h]
  \centering
  \small
  \caption{Accuracy comparison of models using different feature learning methods under various text prompt qualities.
  $\Delta$ denotes the accuracy difference before and after adding the inter-modal alignment loss $\mathcal{L}_{\text{CLIP}}$, and ``$\downarrow$'' indicates accuracy drop.}
  \label{tab:retrieval_noise_acc}
  \begin{tabular}{ccccc|ccc}
    \toprule
    \multirow{2}{*}{Text retrieval models} & \multirow{2}{*}{Prompt qualities} & \multicolumn{3}{c|}{Stanford Cars} & \multicolumn{3}{c}{Flowers102}  \\
    \cmidrule(lr){3-5} \cmidrule(lr){6-8}
    & & $\mathcal{L}_{\text{SSR}^2}$ & $\mathcal{L}_{\text{SSR}^2} + \mathcal{L}_{\text{CLIP}}$ & $\Delta$ & $\mathcal{L}_{\text{SSR}^2}$ & $\mathcal{L}_{\text{SSR}^2} + \mathcal{L}_{\text{CLIP}}$ & $\Delta$ \\
    \midrule
    CLIP-H/14 & High & $89.2$ & $88.1$ & $1.1$ $\downarrow$ & $93.5$ & $92.9$ & $0.6$ $\downarrow$ \\
    CLIP-L/14 & Medium & $87.5$ & $86.1$ & $1.4$ $\downarrow$ & $92.4$ & $91.8$ & $0.6$ $\downarrow$ \\
    CLIP-B/16 & Low & $85.2$ & $84.0$ & $1.2$ $\downarrow$ & $90.1$ & $89.8$ & $0.3$ $\downarrow$ \\
    \bottomrule
  \end{tabular}
\end{table*}

To further verify that extra inter-modal alignment may be not necessary, and jointly optimizing inter-modal alignment and intra-modal alignment significantly impairs feature learning, we validate the role of inter-modal alignment under different text prompt qualities (\ie, which associate with different noise levels). 
Specifically, in the RTA strategy, we employ three frozen CLIP models~\citep{Radford:ICML2021-CLIP}---namely CLIP-H/14, CLIP-L/14, and CLIP-B/16---to perform  prompt searching, in order to evaluate how inter-modal alignment affects model performance under varying prompt qualities.
The results in Table~\ref{tab:retrieval_noise_acc} show %demonstrate 
that, on Stanford Cars and Flowers102, introducing an extra $\mathcal{L}_{\text{CLIP}}$ loss for joint training with $\mathcal{L}_{\text{SSR}^2}$ consistently degrades clustering performance at all levels of prompt quality.

\begin{table*}[h]
\centering
\small
\caption{Comparison to uni-modal counterparts. ACC (\%) on generic and fine-grained datasets is reported.}
\label{tab:unimodal}
\setlength{\tabcolsep}{1.3mm}{
\begin{tabular}{l ccc ccc ccc ccc ccc ccc ccc}
\toprule
& \multicolumn{3}{c}{CIFAR-10} & \multicolumn{3}{c}{CIFAR-100} & \multicolumn{3}{c}{CUB} & \multicolumn{3}{c}{Stanford Cars} & \multicolumn{3}{c}{Oxford Pets} & \multicolumn{3}{c}{Flowers102} \\
\cmidrule(lr){2-4} \cmidrule(lr){5-7} \cmidrule(lr){8-10} \cmidrule(lr){11-13} \cmidrule(lr){14-16} \cmidrule(lr){17-19}
Method & All & Old & New & All & Old & New & All & Old & New & All & Old & New & All & Old & New & All & Old & New \\
\midrule
GCD & 91.5 & \textbf{97.9} & 88.2 & 73.0 & 76.2 & 66.5 & 51.3 & 56.6 & 48.7 & 39.0 & 57.6 & 29.9 & 80.2 & 85.1 & 77.6 & 74.4 & 74.9 & 74.1 \\
\rowcolor{gray!20}
GCD+SSR$^2$ & 92.5 & 96.4 & 91.6 & 73.9 & 79.0 & 63.2 & 51.9 & 55.0 & 47.1 & 47.9 & 56.1 & 47.3 & 83.6 & 87.7 & 79.8 & 80.0 & 83.3 & 78.5 \\
SimGCD & 97.1 & 95.1 & \textbf{98.1} & 80.1 & 81.2 & 77.8 & 60.3 & \textbf{65.6} & 57.7 & 53.8 & \textbf{71.9} & 45.0 & 87.7 & 85.9 & 88.6 & 71.3 & 80.9 & 66.5 \\
\rowcolor{gray!20}
SimGCD+SSR$^2$ & \textbf{97.6} & 97.5 & 97.7 & \textbf{81.1} & \textbf{82.5} & \textbf{78.9} & \textbf{60.8} & 64.7 & \textbf{59.0} & \textbf{57.1} & 66.8 & \textbf{53.9} & \textbf{90.0} & \textbf{89.8} & \textbf{91.2} & \textbf{81.6} & \textbf{83.5} & \textbf{80.1} \\
\bottomrule
\end{tabular}
}

\end{table*}

\myparagraph{Evaluation on Uni-Modal Representation Learning}
To evaluate its effectiveness of the proposed $\mathcal{L}_\mathrm{SSR^2}$ as an intra-modal alignment loss,
we apply $\mathcal{L}_\mathrm{SSR^2}$ to uni-modal GCD counterparts and report their clustering performance.
Specifically, for GCD and SimGCD frameworks, we keep their pre-trained models and clustering algorithms unchanged, but %and 
replace the supervised and unsupervised contrastive loss $\mathcal{L}_\mathrm{con}$ with our proposed $\mathcal{L}_\mathrm{SSR^2}$.
As can be read from Table~\ref{tab:unimodal} that, using our $\mathcal{L}_\mathrm{SSR^2}$ for representation learning achieves improvements on most cases.  %all 
% test datasets. %while achieves a relatively less significant improvement on CUB.

\begin{figure}[h]
    \centering
    \includegraphics[width=0.9\linewidth]{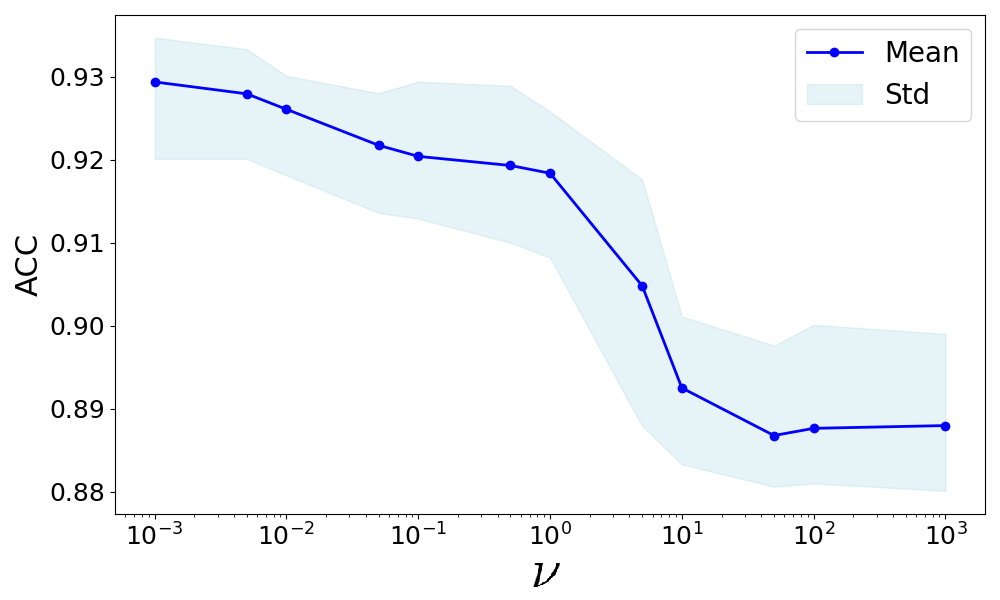}
    \caption{Mean accuracy with standard deviation over %and standard accuracy (
    3 trials of ``All'' categories with varying penalty weight $\nu$ on Flowers102. }
    \label{fig:ablation_comb_varphi}
\end{figure}

% lambda, gamma, mu and alpha are used in Eqs (2), (3) and (6) in the manuscript. Thus varphi is used here
\myparagraph{Evaluation on %the Role 
Effect of Inter-Modal Alignment against Intra-Modal Alignment}
To evaluate the effect %role 
of performing inter-modal alignment against %and 
intra-modal alignment, %interactions, 
we combine our proposed intra-modal alignment loss $\mathcal{L}_\text{SSR$^2$}$ with the inter-modal alignment loss $\mathcal{L}_\text{CLIP}$ by introducing a tradeoff parameter $\nu >0$, \ie,
$$\mathcal{L}_\text{SSR$^2$} +\nu \cdot \mathcal{L}_\text{CLIP},$$
where using a larger $\nu$ means to %is to a tpenalty parameter weight of 
emphasizing more on the inter-model alignment. % via $\mathcal{L}_\text{CLIP}$.
In Figure~\ref{fig:ablation_comb_varphi}, we report the accuracy as a function with respect to a varying %penalty weight 
$\nu$ on Flowers102. % dataset.
Existing multi-modal GCD frameworks, such as GET~\citep{Wang:CVPR25-get}, assume that %the learning of 
enforcing an inter-modal alignment does not affect that of intra-modal alignment, and %naively 
simply treat these two alignments %learning processes 
in representation learning as independent and equally important.
However, our experiments reveal here that the more emphasizing upon inter-modal alignment (\ie, via $\mathcal{L}_\text{CLIP}$ with a larger % by increasing 
$\nu$), rather than upon the intra-modality alignment (\ie, via $\mathcal{L}_{\text{SSR}^2}$), the lower the clustering accuracy is.
% we can see from Figure~\ref{fig:ablation_comb_varphi} that the learning of inter-modal alignment significantly impairs the learning of intra-modal alignment as $\nu$ increases.

\myparagraph{More Results on Consistency Measure $\rho$}
Recall that we define a consistency measure $\rho$ in Eq.(\ref{eq:compactness-vs-cut}) %by the ratio of the weights on edges $\rho$ to quantize 
to quantify the intra-modal consistency of the embeddings to explain why extra inter-modal alignment could %can 
be unnecessary. 
Here, we additionally report the consistency measure $\rho$ when training via inter-modal alignment loss (\ie, $\mathcal{L}_\text{CLIP}$), intra-modal alignment losses (\ie, $\mathcal{L}_\text{con}$) and their combinations (\ie, $\mathcal{L}_\text{con}+\mathcal{L}_\text{CLIP}$) on Stanford Cars.
We display the experimental results in Figure~\ref{Fig:re_con}.
As can
As can be seen that, the inter-modal alignment also damages the learning of other intra-modal losses such as the widely adopted supervised and unsupervised contrastive loss $\mathcal{L}_\mathrm{con}$.
%
% This confirms that the aforementioned trend is not exclusive to scenarios where $\mathcal{L}_\mathrm{CLIP}$ is combined with our proposed $\mathcal{L}_\mathrm{SSR^2}$.

\begin{figure}[h]
    \centering
    \begin{subfigure}[b]{0.5\linewidth}
           \includegraphics[width=\textwidth]{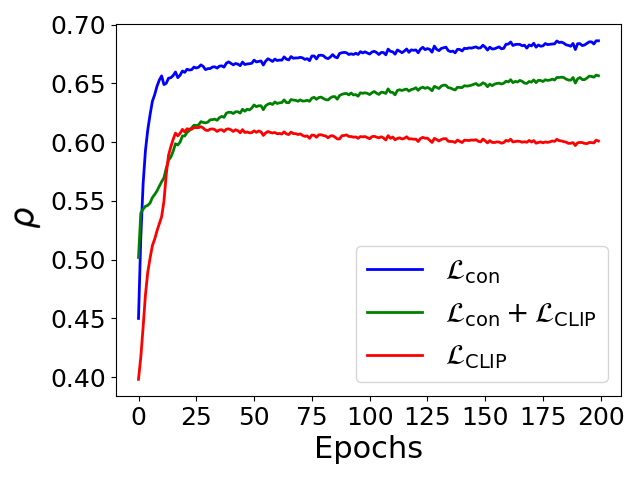}
            \caption{Image embeddings}
    \end{subfigure}\hfill
    \begin{subfigure}[b]{0.5\linewidth}
           \includegraphics[width=\textwidth]{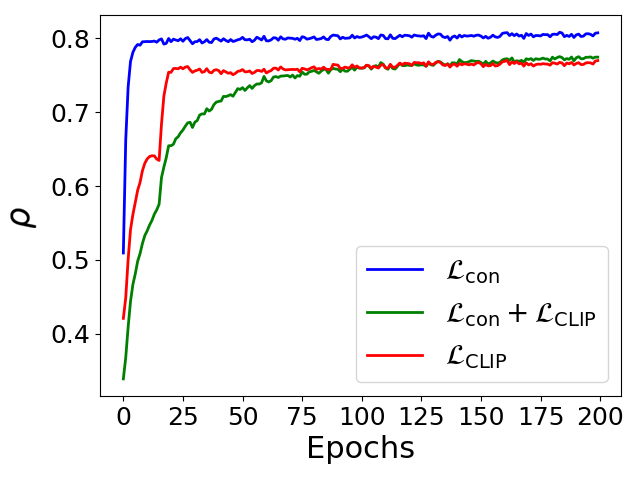}
            \caption{Text embeddings}
    \end{subfigure}
\caption{{Consistency measure $\rho$ %Ratio curves 
as a function of training epoch under different losses on Stanford Cars.}}
\label{Fig:re_con}
\end{figure}

\begin{figure*}[h]
    \centering
    \includegraphics[width=0.95\linewidth]{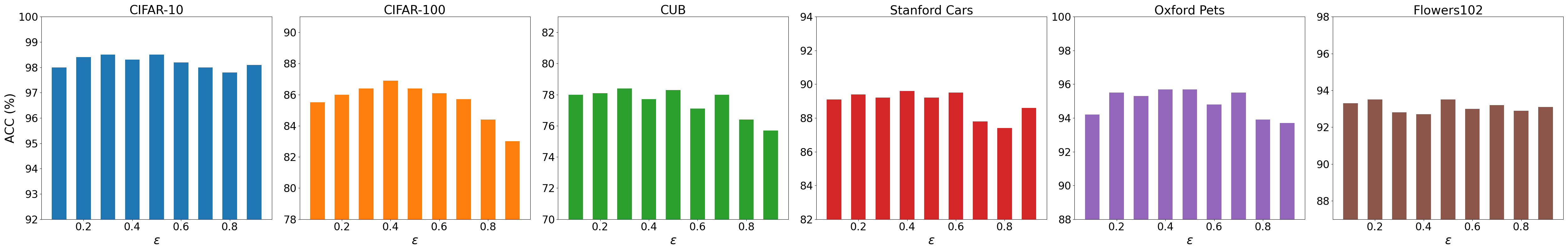}
    \caption{ACC (\%) with varying $\epsilon$ in SSR$^2$ across six benchmark datasets.}
    \label{fig:eps}
\end{figure*}

\subsection{Evaluation on Parameter $\epsilon$ in SSR$^2$}
To evaluate the impact of using different hyper-parameters $\epsilon$ in our SSR$^2$-GCD, we keep %set 
all other hyper-parameters fixed, and conduct experiments under %via using 
different parameter $\epsilon$ on six benchmark datasets, \ie, CIFAR-10, CIFAR-100, CUB, Stanford Cars, Oxford Pets and Flowers102.
Experimental results are reported in Figure~\ref{fig:eps}.
As can be seen, our SSR$^2$-GCD framework is not sensitive to $\epsilon$ and achieves the best performance when $\epsilon$ is in the range of $[0.2, 0.5]$.

\begin{figure*}[h]
    \centering
    \begin{subfigure}[b]{0.3\linewidth}
           \includegraphics[width=\textwidth]{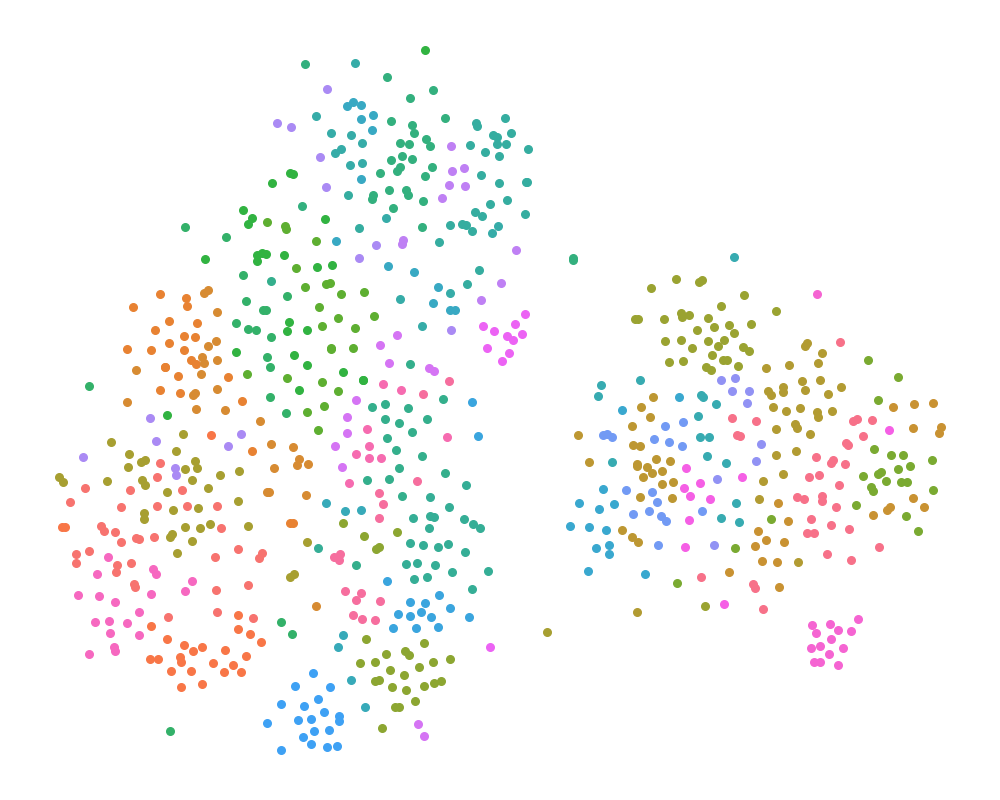}
            \caption{Using a frozen CLIP}
            \label{Fig:(a)_in_vis}
    \end{subfigure}\hfill
    \begin{subfigure}[b]{0.3\linewidth}
           \includegraphics[width=\textwidth]{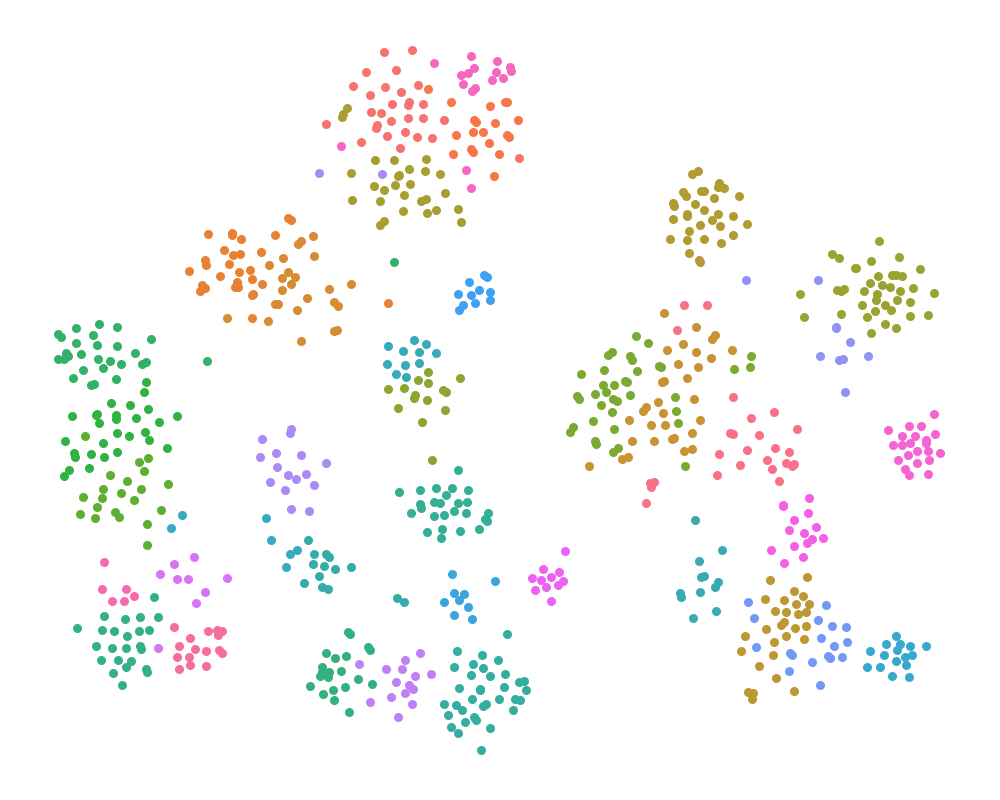}
            \caption{Trained via $\mathcal{L}_\mathrm{CLIP}$}
            \label{Fig:(b)_in_vis}
    \end{subfigure}\hfill
    \begin{subfigure}[b]{0.3\linewidth}
           \includegraphics[width=\textwidth]{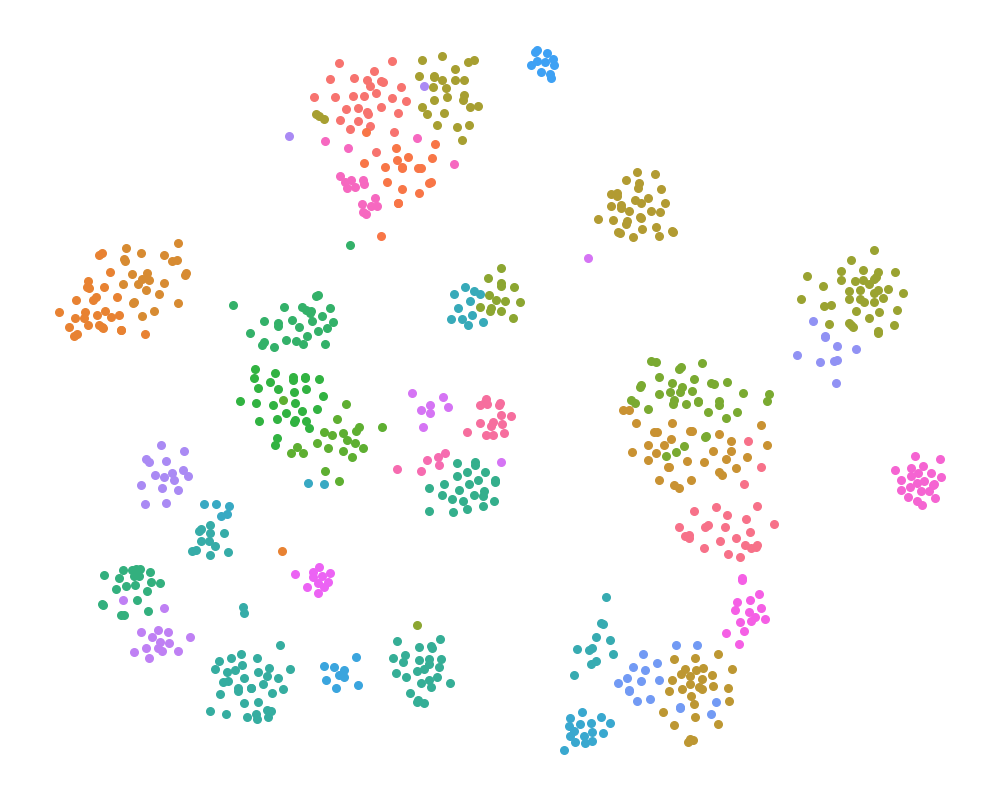}
            \caption{Trained via $\mathcal{L}_\mathrm{SSR^2}$}
    \end{subfigure}\hfill
    \begin{subfigure}[b]{0.3\linewidth}
           \includegraphics[width=\textwidth]{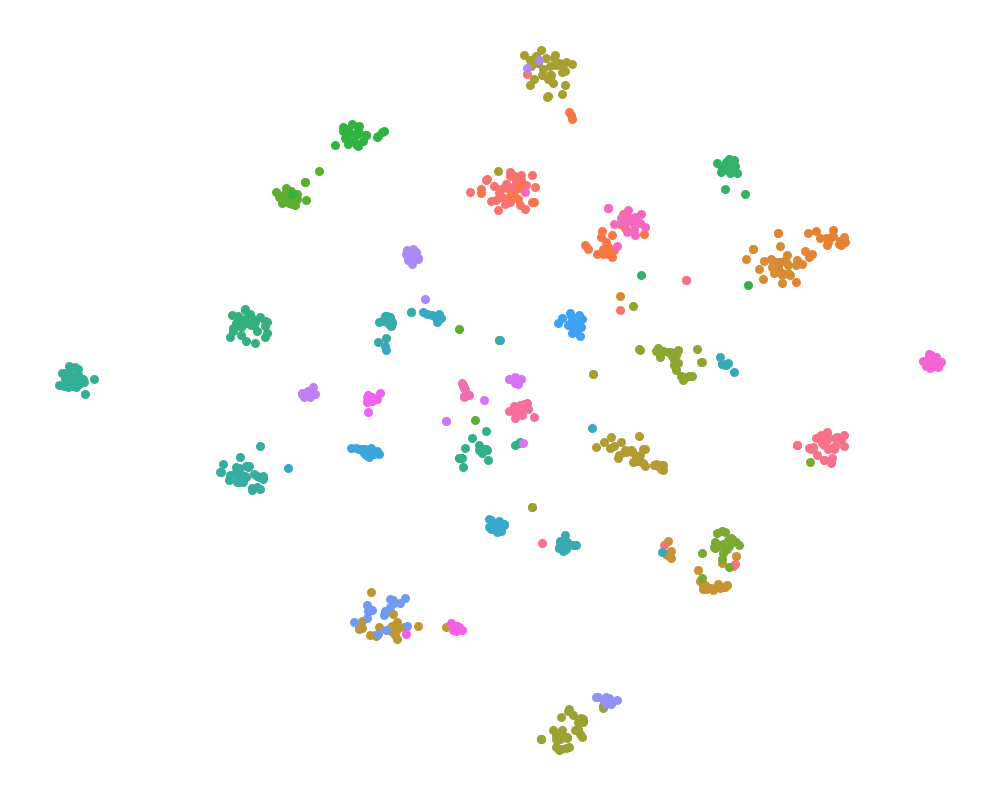}
            \caption{Using a frozen CLIP}
    \end{subfigure}\hfill
    \begin{subfigure}[b]{0.3\linewidth}
           \includegraphics[width=\textwidth]{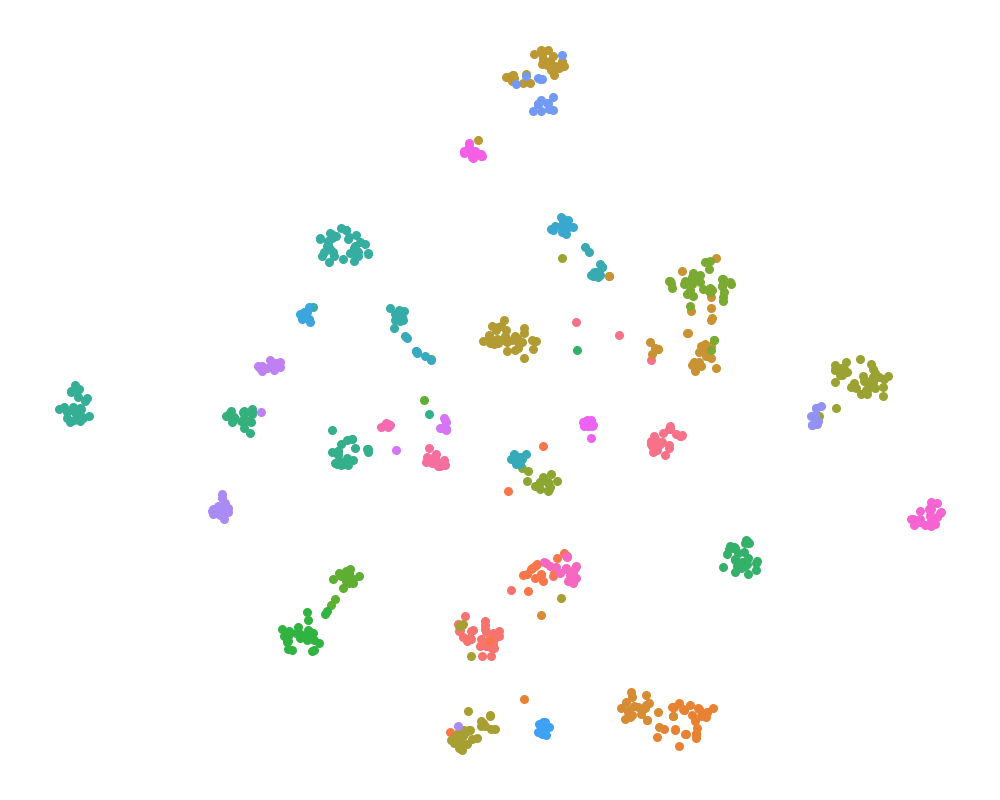}
            \caption{Trained via $\mathcal{L}_\mathrm{CLIP}$}
    \end{subfigure}\hfill
    \begin{subfigure}[b]{0.3\linewidth}
           \includegraphics[width=\textwidth]{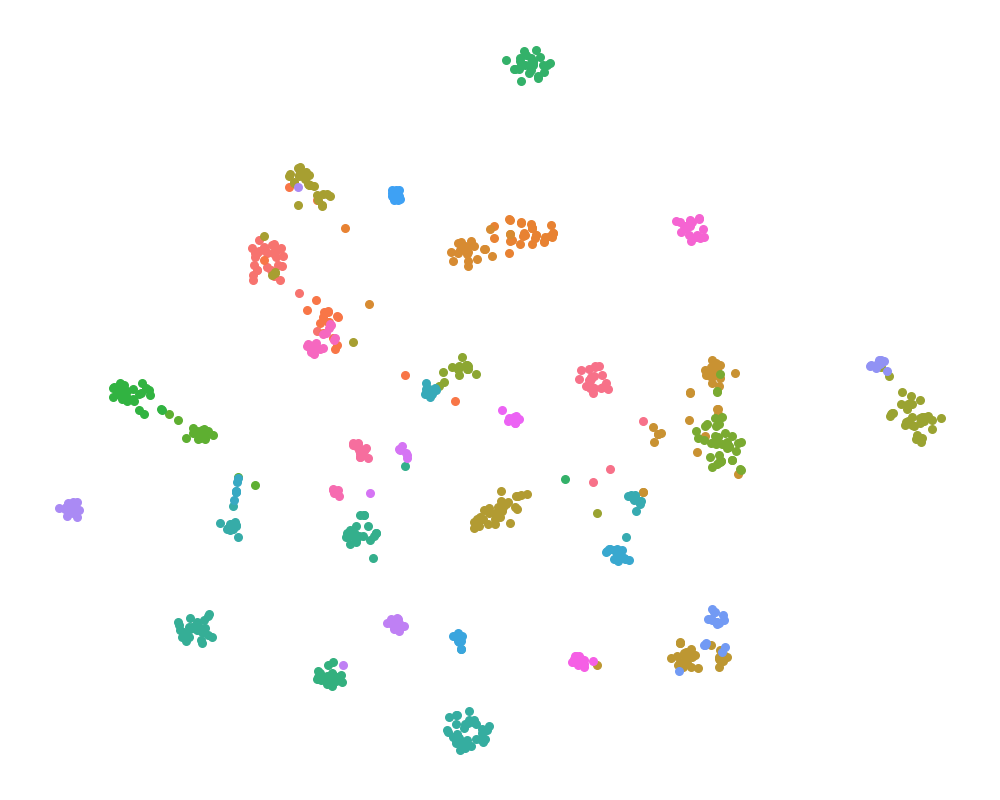}
            \caption{Trained via $\mathcal{L}_\mathrm{SSR^2}$}
    \end{subfigure}
\caption{Visualization of image embeddings (\textbf{top}) and text embeddings (\textbf{bottom}) on Oxford Pets.}
\label{Fig:vis_pets}
\end{figure*}

\subsection{Visualization}
In Figure~\ref{Fig:vis_pets}, we use $t$-SNE to visualize the image embeddings and text embeddings of our SSR$^2$-GCD framework using different representation learning methods on Oxford Pet, which comprises 12 categories of cats and 25 categories of dogs.
As can be seen, the image embeddings produced by using the frozen CLIP image encoder exhibit a distribution that can only be roughly partitioned into two classes (i.e., ``cat" and ``dog").
In contrast, thanks to our proposed RTA strategy, the distribution of text embeddings produced by using the frozen CLIP text encoder assisted with our RTA exhibits significant discriminability.
%
% Such discrepancies between the two modalities may explain why optimizing the inter-modal alignment loss can enhance intra-class discriminability to some extent---the learning of image embeddings is largely guided by the inherently discriminative text embeddings.
%
Meanwhile, training via $\mathcal{L}_\text{CLIP}$ fails to learn well-aligned intra-modal relationships (See, \eg, Figure~\ref{Fig:(a)_in_vis} and Figure~\ref{Fig:(b)_in_vis} for comparison).
In contrast, our SSR$^2$-GCD learns discriminative and well-balanced representations for both modalities.

\begin{figure}[h]
    \centering
    \begin{subfigure}[b]{0.45\linewidth}
           \includegraphics[width=\textwidth]{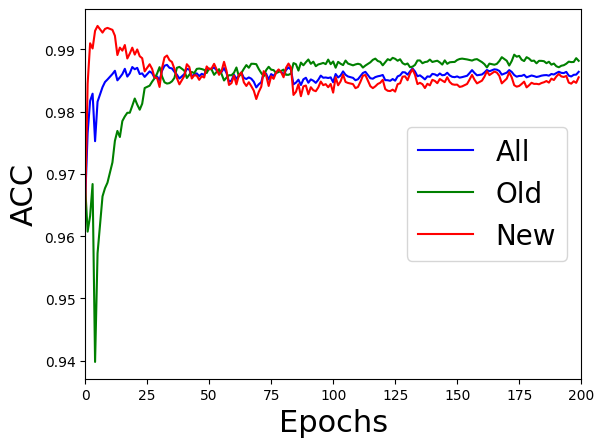}
            \caption{CIFAR-10}
    \end{subfigure}\hfill
    \begin{subfigure}[b]{0.45\linewidth}
           \includegraphics[width=\textwidth]{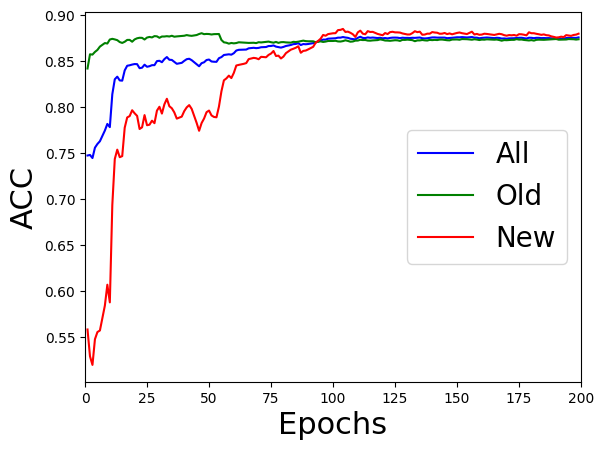}
            \caption{CIFAR-100}
    \end{subfigure}\hfill
    \begin{subfigure}[b]{0.45\linewidth}
           \includegraphics[width=\textwidth]{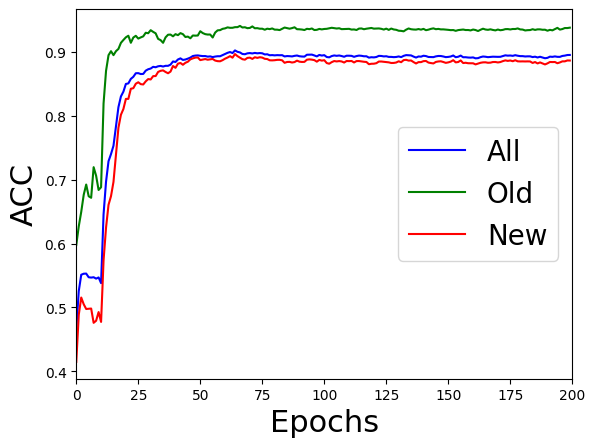}
            \caption{Stanford Cars}
    \end{subfigure}\hfill
    \begin{subfigure}[b]{0.45\linewidth}
           \includegraphics[width=\textwidth]{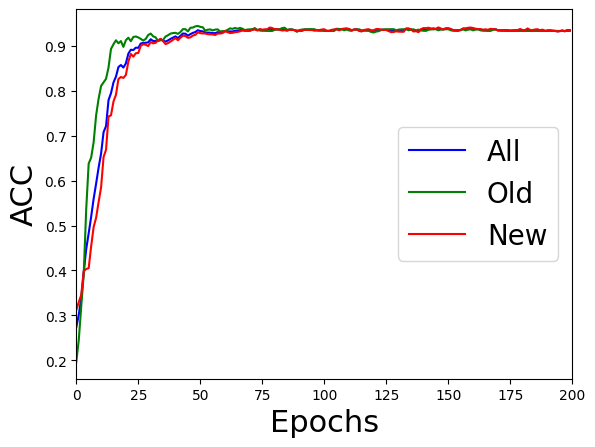}
            \caption{Flowers102}
    \end{subfigure}
\caption{\text{ACC curves on benchmark datasets.}}
\label{Fig:acc_curve}
\end{figure}

\subsection{Learning Curves}
We compute the clustering accuracy (ACC) as a function of epoch during training period, and display the ACC curves on different datasets in Figure~\ref{Fig:acc_curve}. 
We can observe that our SSR$^2$-GCD converges and achieves stable clustering performance on ``All'' categories roughly within 50 epochs.

\begin{table}[h]
\centering
\semismall
\caption{Running time and memory costs on Flowers102. ``$\dagger$'' denotes to use the method in TextGCD~\citep{Zheng:ECCV24-textgcd} to produce text features.}
    \setlength{\tabcolsep}{1.3mm}
    \begin{tabular}{lccccc}
    \hline
    \multirow{2}{*}{Methods} & \multicolumn{3}{c}{Running Time (sec/iter)} & \multicolumn{2}{c}{Memory (MB)} \\
    \cmidrule(lr){2-4} \cmidrule(lr){5-6}
     & Forward & Backward & Overall & w/o RTA$^\dagger$ & w/ RTA \\
     \hline
    TextGCD~\citep{Zheng:ECCV24-textgcd} & $0.34$ & $6.7\times 10^{-3}$ & $0.69$ & 7,622 & - \\
    SSR$^2$-GCD & $0.57$ & $7.5\times 10^{-2}$ & $0.92$ & 8,035 & 11,080 \\
    \hline
    \end{tabular}
    \label{tab:cost}
\end{table}

\subsection{Computation Time and Memory Costs}
We report the running time and memory costs on Flowers102 of our SSR$^2$-GCD %is reported 
in Table~\ref{tab:cost}.
Note that the coding rate term $\log\det(\cdot)$ is cheap to handle because its scale is kept to $\min\{|B|,d\}$ due to the fact that $\log\det\left(\mathbf{I} + \mathbf{Z}\mathbf{Z}^\top\right) = \log\det\left(\mathbf{I} + \mathbf{Z}^\top\mathbf{Z}\right)$.
Comparing to TextGCD~\cite{Zheng:ECCV24-textgcd}, the increased costs of our SSR$^2$-GCD is due to the RTA step in forward pass.

\subsection{Performance Evaluation on Using Varying $K$} 
To evaluate the sensitivity of our SSR$^2$-GCD to the category number $K$, %mis-specification, 
we change %set 
the output dimensions of the classifiers $d_\text{cls}$ to deviate from the total number of categories $K$, and report the clustering accuracy (ACC), Normalized Mutual Information (NMI) and Adjusted Rand Index (ARI) of our SSR$^2$-GCD with varying $d_\text{cls}$ on Flowers102 in Figure~\ref{fig:k_misspecific}.
As can be observed that, our SSR$^2$-GCD achieves relatively high clustering performance when $d_\text{cls} \approx K$ (\eg, $d_\text{cls}\in\{100,102,104\}$). Specifically, our SSR$^2$-GCD is sensitive to under-specification, where all metrics degrade sharply when $d_\text{cls} < K$; whereas our SSR$^2$-GCD exhibits good robustness to over-specification, where all metrics degrades slightly. We notice of that NMI remains nearly unchanged as $d_\text{cls}$ increases beyond $K$. 

\begin{figure}
    \centering
    \includegraphics[width=0.95\linewidth]{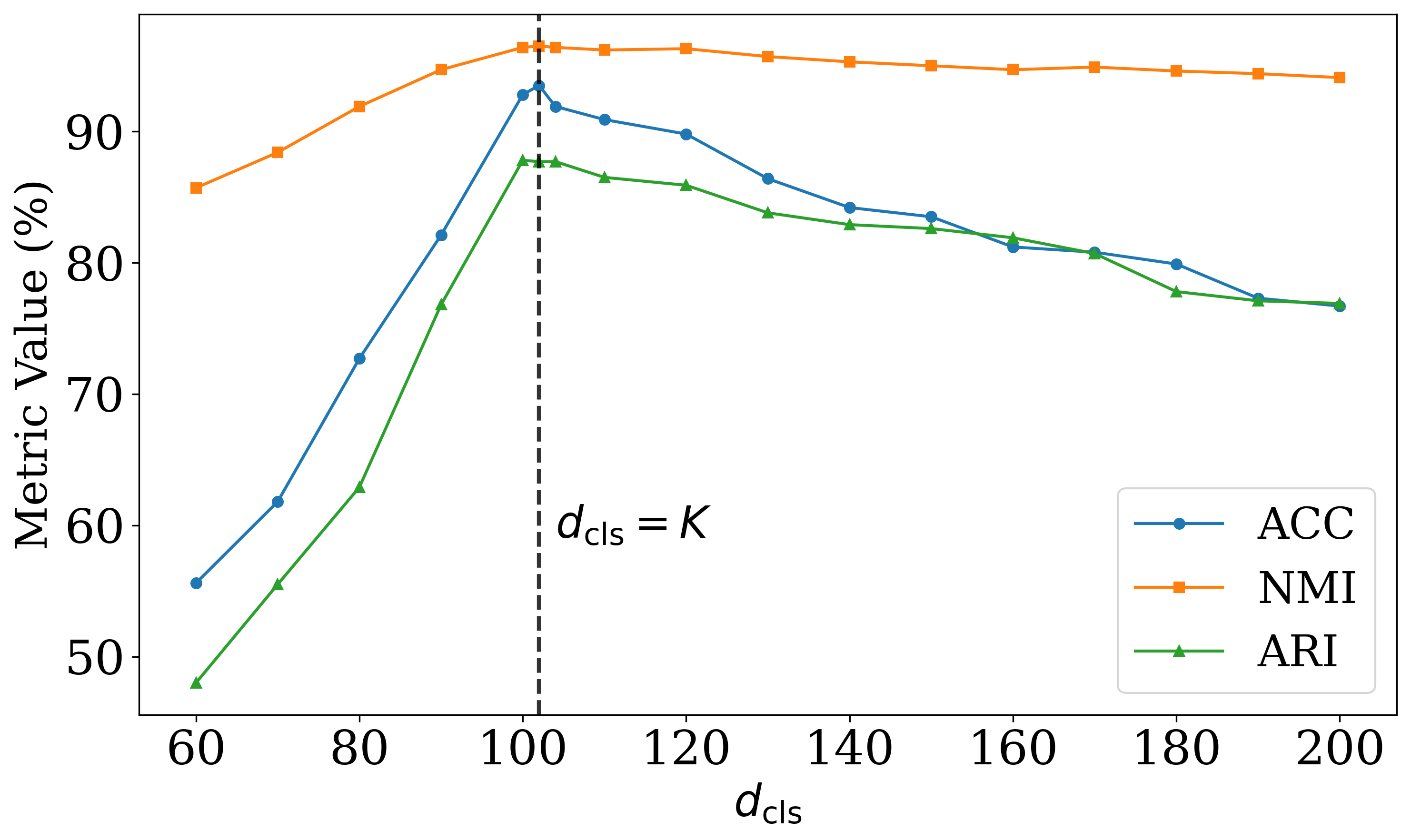}
    \caption{Evaluation on Clustering performance ACC, NMI and ARI of our SSR$^2$-GCD with varying $d_\text{cls}$ on Flowers102.}
    \label{fig:k_misspecific}
\end{figure}

}
% End: appendix

\end{document}